\documentclass[conference]{IEEEtran}
\IEEEoverridecommandlockouts

\usepackage{cite}
\usepackage{amsmath,amssymb,amsfonts}
\usepackage{algorithmic}
\usepackage{graphicx}
\usepackage{textcomp}
\usepackage{xcolor}

\usepackage[table,dvipsnames]{xcolor}
\usepackage{booktabs}
\usepackage{wrapfig}
\usepackage{amsmath}
\usepackage{subcaption}
\usepackage{multirow}
\usepackage{bbm}
\usepackage[ruled,vlined]{algorithm2e}
\usepackage{pifont}
\usepackage[normalem]{ulem}
\usepackage{stfloats}

\def\BibTeX{{\rm B\kern-.05em{\sc i\kern-.025em b}\kern-.08em
    T\kern-.1667em\lower.7ex\hbox{E}\kern-.125emX}}
\begin{document}

\title{CATCHFed: Efficient Unlabeled Data Utilization for Semi-Supervised Federated Learning in Limited Labels Environments
}



\author{
\IEEEauthorblockN{Byoungjun Park\IEEEauthorrefmark{1}, Pedro Porto Buarque de Gusmão\IEEEauthorrefmark{2}, Dongjin Ji\IEEEauthorrefmark{3}, Minhoe Kim\IEEEauthorrefmark{4}}

\IEEEauthorblockA{\IEEEauthorrefmark{1}Dept. of Computer and Information Science, Korea University, Sejong, South Korea}
\IEEEauthorblockA{\IEEEauthorrefmark{2}Department of Computer Science, University of Surrey, Guildford, United Kingdom}
\IEEEauthorblockA{\IEEEauthorrefmark{3}Dept. of Semiconductor Engineering, Seoul National University of Science and Technology, Seoul, South Korea}
\IEEEauthorblockA{\IEEEauthorrefmark{4}Dept. of Electrical and Information Engineering, Seoul National University of Science and Technology, Seoul, South Korea}

\vspace{0.5\baselineskip} 
\IEEEauthorblockA{
pbjun3348@korea.ac.kr\IEEEauthorrefmark{1},
p.gusmao@surrey.ac.uk\IEEEauthorrefmark{2},
jdj0524@seoultech.ac.kr\IEEEauthorrefmark{3},
kimminhoe@seoultech.ac.kr\IEEEauthorrefmark{4}
}
}

\maketitle

\begin{abstract}
Federated learning is a promising paradigm that utilizes distributed client resources while preserving data privacy. Most existing FL approaches assume clients possess labeled data, however, in real-world scenarios, client-side labels are often unavailable. Semi-supervised Federated learning, where only the server holds labeled data, addresses this issue. However, it experiences significant performance degradation as the number of labeled data decreases. 
To tackle this problem, we propose \textit{CATCHFed}, which introduces client-aware adaptive thresholds considering class difficulty, hybrid thresholds to enhance pseudo-label quality, and utilizes unpseudo-labeled data for consistency regularization. 
Extensive experiments across various datasets and configurations demonstrate that CATCHFed effectively leverages unlabeled client data, achieving superior performance even in extremely limited-label settings.
\end{abstract}


\section{Introduction} \label{sec:intro}
Due to recent advances in artificial intelligence (AI) technology, various fields are benefiting from the results of extensive research and development.
Unlike other technologies, training data is an essential component of AI which includes both labeled and unlabeled data. 
Especially, labeled data (size, quality) can greatly affect the performance of supervised learning-based models.
Although labeled data are seldom abundant in real-world settings, many of the works in the literature are usually based on the implicit assumption that such labels are ubiquitously available.
Except for a few specific tasks, datasets with sufficiently rich labels are seldom available.
To address this, Semi-Supervised Learning uses unlabeled data as a source of information along with the labeled data.

Another problem of training data is accessibility of data.
In many cases, training data are gathered from various distributed nodes, which might contain privacy-sensitive data. To address this, Federated Learning (FL) leverages distributed data from numerous clients while offering strong privacy protection \cite{mcmahan2017communication}. Specifically, each client trains a model using only its local data, then transmits only the learned model parameters (or gradients) to the server to update the global model. This training method helps protect data privacy while still making good use of diverse data.
However, most FL studies assume that clients possess labeled data or that labels are naturally generated through user interactions \cite{li2020federated, karimireddy2020scaffold, wu2021fast, rehman2023dawa}, while in practice, labeling is often difficult or even infeasible \cite{jin2023federated}.
For instance, consider training an image-recognition module for autonomous vehicles with deep neural networks. The images captured on each vehicle frequently contain privacy-sensitive information, e.g., pedestrians, license plates, and similar details, that must remain local to the device. As a result, the data cannot be forwarded to a central server and lacks labels.

\begin{figure}[t]
\vspace{-5pt}
  \centering
  \includegraphics[width=\linewidth]{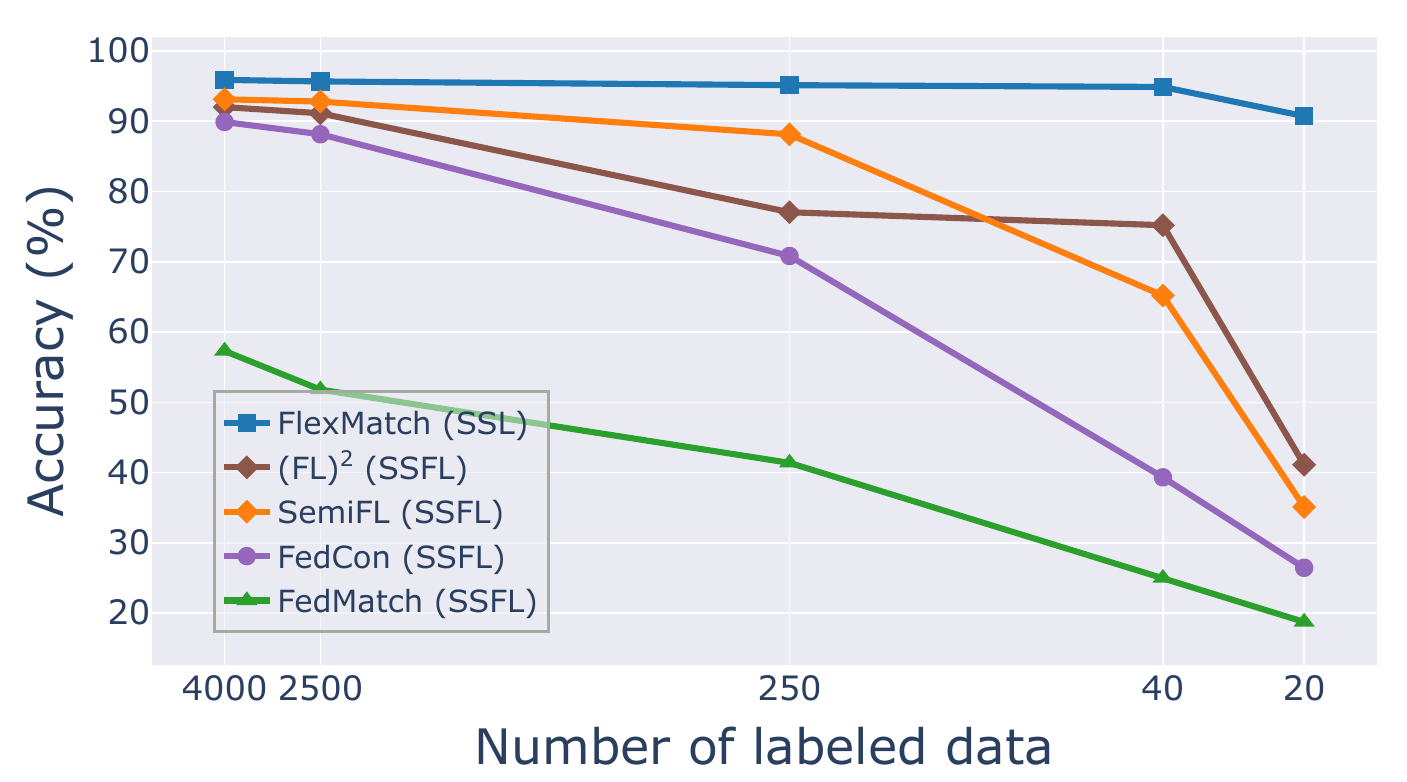}
  \caption{CIFAR-10 accuracy comparison by labeled sample count (FlexMatch\cite{sohn2020fixmatch}, $(FL)^2$\cite{lee20242}, SemiFL\cite{diao2022semifl}, FedMatch\cite{jeong2020federated}, and FedCon\cite{long2021fedcon})}
  \label{fig:ssl_vs_ssfl_test_acc}
\end{figure}

\begin{figure}[t]
\vspace{-5pt}
  \centering
  \begin{subfigure}[t]{0.49\linewidth}
    \includegraphics[width=\linewidth]{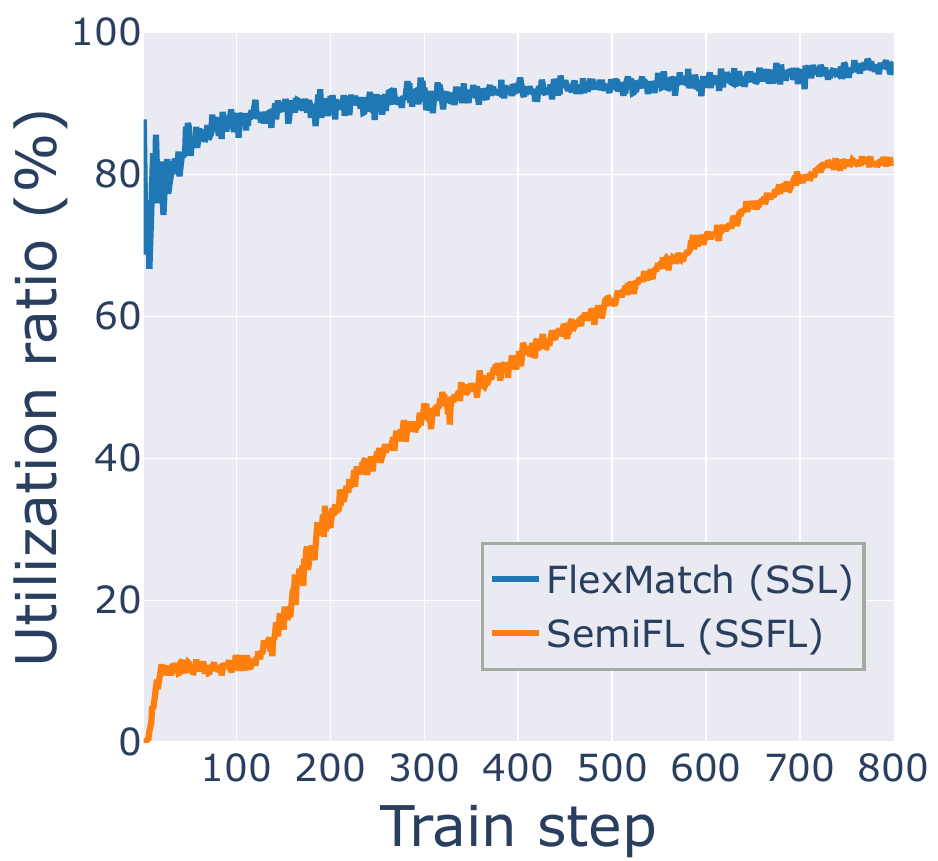}
    \caption{Utilization ratio}
    \label{fig:ssl_vs_ssfl_util_ratio}
  \end{subfigure}\hfill
  \begin{subfigure}[t]{0.49\linewidth}
    \includegraphics[width=\linewidth]{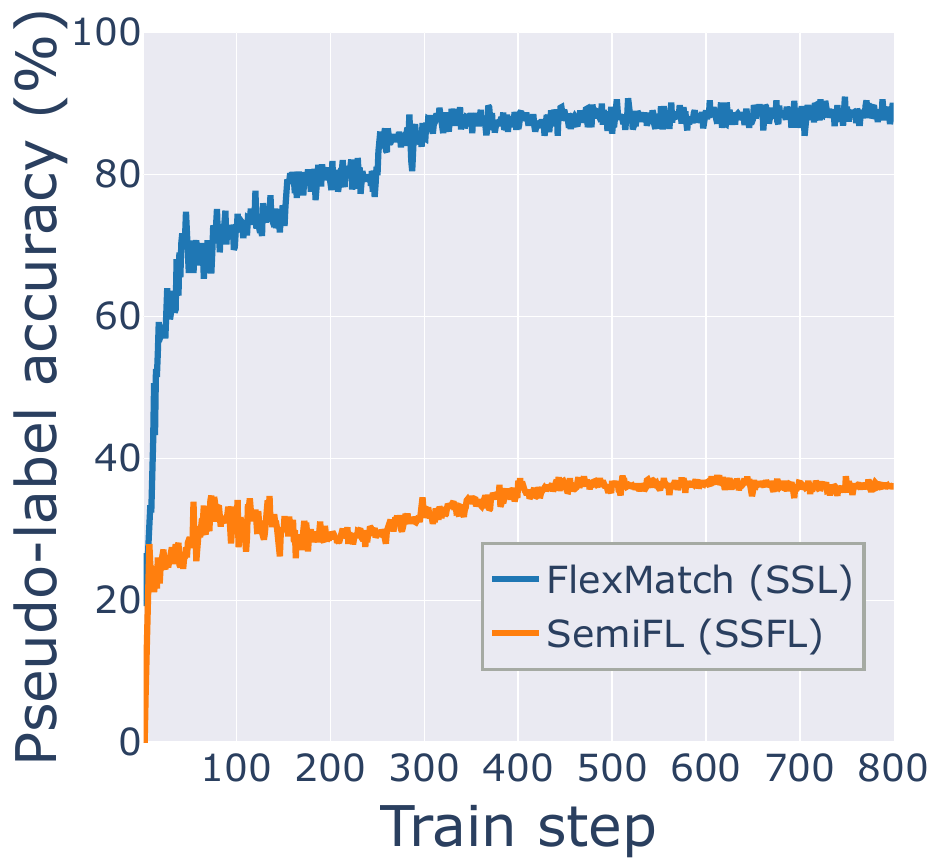}
    \caption{Pseudo-label accuracy}
    \label{fig:ssl_vs_ssfl_pl_acc}
  \end{subfigure}
  \caption{Comparison of utilization ratio and pseudo-label accuracy between FlexMatch\cite{zhang2021flexmatch} and SemiFL\cite{diao2022semifl} on CIFAR-10 with 20 labeled samples.}
  \label{fig:ssl_ssfl_pl_ratio_acc}
  \vspace{-10pt}
\end{figure}

To address this challenge, Semi-Supervised Federated Learning (SSFL) has been proposed, which combines Semi-Supervised Learning (SSL) with FL. While various scenarios exist, in this paper, we assume a setting where labeled data exists only on the server, and clients hold a large amount of unlabeled data \cite{zhang2021improving, long2021fedcon, jeong2020federated, diao2022semifl, lee20242}. Recent SSL studies have shown that even with a small labeled dataset, performance can approach or surpass that of fully supervised learning \cite{berthelot2019mixmatch, sohn2020fixmatch, zhang2021flexmatch, nguyen2024sequencematch, yu2023inpl}. 
SSFL methods based on SSL also perform well when the amount of labeled data is sufficient. 
However, the main challenge of SSFL is that performance degrades significantly in label-scarce settings, as shown in Fig. \ref{fig:ssl_vs_ssfl_test_acc}. 
The performance gap between SSL and SSFL is much larger in label-scarce settings.
We identify the following three factors as primary reasons for this performance degradation: 1) confirmation bias stemming from reduced labeled data in a setting where labeled and unlabeled data are separated, 2) inefficient and limited use of unlabeled data, and 3) low quality of pseudo-labels. Unlike in centralized SSL, SSFL clients train using only unlabeled data. This increases the risk of confirmation bias, where incorrect pseudo-labels are treated as ground truth. Furthermore, SSFL tends to utilize fewer unlabeled samples, such as via pseudo-labeling, compared to SSL, and the quality of pseudo-labels is generally lower, as shown in Fig. \ref{fig:ssl_ssfl_pl_ratio_acc}.

Our main contributions toward overcoming these limitations are as follows:

\begin{itemize}
 \item We introduce an adaptive threshold that reflects the difficulty of the class for each client, improving the utilization of unlabeled data. To mitigate early confirmation bias, we incorporate a warm-up phase at the beginning of training during which most unlabeled data are used.

 \item We enhance the traditional confidence-based thresholding by introducing an additional energy-based threshold, enabling stricter selection and more accurate pseudo-labels. The enhanced quality of pseudo-labels directly translates into higher model performance.

 \item We incorporate unpseudo-labeled data that would otherwise be filtered out via hybrid thresholding into training through consistency regularization instead of discarding them, making better use of unpseudo-labeled data and improving generalization. 
\end{itemize}

\section{Related works and background} \label{sec:related_work}

\subsection{Semi-supervised learning} \label{subsec:ssl}
Semi-Supervised Learning trains a model using both labeled and unlabeled data, assuming that the amount of unlabeled data is much larger than the labeled portion. SSL revolves around two main ideas: \textit{entropy minimization} and \textit{consistency regularization}. Entropy minimization aims to reduce uncertainty by making the model's predictions more confident. Notable examples  include temperature scaling\cite{bachman2014learning}, which sharpens the model's predictions, and pseudo-labeling\cite{lee2013pseudo}, which assigns temporary labels. On the other hand, consistency regularization assumes that the model's predictions should remain stable even when noise is applied to the data. To achieve this, the model is trained so that predictions for various augmented versions of the same input closely match those for the original input. Representative SSL methods combine the two approaches mentioned above in various ways. For example, FixMatch \cite{sohn2020fixmatch} generates pseudo-labels for weakly augmented images when confidence is high, and applies consistency regularization by training strongly augmented versions to match them. FlexMatch \cite{zhang2021flexmatch} adjusts thresholds per class based on learning difficulty to improve pseudo-labeling. SequenceMatch \cite{nguyen2024sequencematch} includes low-confidence samples in consistency regularization and uses medium augmentations to reduce confirmation bias. InPL \cite{yu2023inpl} utilizes energy scores instead of softmax-based confidence to generate pseudo-labels.

\subsection{Semi-supervised federated learning} \label{subsec:ssfl}
Semi-supervised federated learning combines FL and SSL to enable effective learning in limited-label scenarios. While various settings are exist \cite{liu2021federated, yang2021federated, liang2022rscfed, li2023class}, this paper focuses on the label-at-server scenario, where labeled data exists only on the server and clients hold only unlabeled data. FedMatch \cite{jeong2020federated} not only combines FL with SSL methods such as FixMatch but also introduces an inter-client consistency loss term to maintain prediction consistency across clients. Moreover, it separates the server and client parameters, mitigating performance degradation caused by unlabeled data. FedCon \cite{long2021fedcon} argues that simply combining FL with SSL methods, such as FedMatch leads to performance degradation and proposes a new SSFL framework integrating contrastive learning. SemiFL \cite{diao2022semifl} modifies the usual SSFL training so that clients generate pseudo-labels only once using the global model, then conduct local training based on those labels. After aggregation, the resulting global model is further fine-tuned with the server’s labeled data, an approach referred to as \textit{alternative training}. With these structural changes, SemiFL \cite{diao2022semifl} can achieve performance comparable to centralized SSL when there is sufficient labeled data on the server. $(FL)^2$ \cite{lee20242} introduces a pseudo-labeling strategy that considers the learning state based on the average confidence of each client, and enforces consistency between the outputs of the original model and its perturbed version on high-confidence data. Through this approach, it demonstrates additional performance gains even in environments with limited labeled data on the server.

However, while centralized SSL maintains high performance even with very limited labeled data, SSFL methods still show a significant drop in accuracy. We attribute this to existing methods not fully utilizing unlabeled client data and failing to resolve confirmation bias. We address this by making better use of clients' unlabeled data and showing that this improves overall performance.

\subsection{Energy score} \label{subsec:energy_score}
Energy score \cite{lecun2006tutorial} is a quantitative measure derived from model logits that reflects how closely an input data matches the training distribution. This allows for the evaluation of how well the input data conforms to the distribution used during training. Specifically, the energy score is defined as follows:

\begin{equation} \label{eq:energy_score}
    E(f^{w}(x)) = -T \cdot \log(\sum_{i=1}^{K} \exp(\frac{f^{w}_i(x)}{T})).
\end{equation}

Here, $f^w_i(\cdot)$ denotes the logits for the $i$th class for input $x$ given model parameters $w$, $K$ is the total number of classes, and $T$ is a hyperparameter known as the temperature. The setting of $T$ regulates the rate of change in the energy score, and it is generally set to $T=1$. \cite{liu2020energy} has demonstrated that the energy score is effectively applied in the field of OOD (out-of-distribution) detection. A low energy score indicates that the data sample is close to the training data distribution (in-distribution), whereas a high energy score suggests that the data is far from the training data distribution.

\section{System model} \label{sec:system_model}
In this section, we describe the SSFL training process and the energy score. The overall SSFL training framework is based on the SemiFL \cite{diao2022semifl} method. We consider label-at-server SSFL where labeled and unlabeled data are separate and cannot be accessed simultaneously during training unlike SSL. As a result, the server uses only labeled data for training, while the clients use only unlabeled data.

\subsection{Server} \label{subsec:server}
On the server side, there is a small set of labeled data $\mathcal{D}^s_l = \{(x^s_n, y^s_n)\}^{D^s_l}_{n=1}$ where $x^s_n$ and $y^s_n$ represent the $n$-th input features and the corresponding labels, respectively, and $D^s_l$ denotes the total number of labeled data on the server (i.e., the size of $\mathcal{D}^s_l$). The superscript $s$ means its presence is on the server. Following SSL practices, a weak augmentation is applied to the input data, which differs from standard supervised training. At communication round $r$, the server-side training loss is defined as follows, based on which the model parameters are updated:

\begin{equation} \label{eq:l_loss}
  L^{s}_{l} = \frac{1}{B^{s}} \sum_{(x^{s}_{b},\,y^{s}_{b}) \in \mathcal{B}^{s}} H\!\bigl(y^{s}_{b}, p_{w^{s}}(y \mid \mathcal{A}_{w}(x^{s}_{b}))),
\end{equation}

Here $\mathcal{B}^s = \{(x^s_b, y^s_b)\}^{B^s}_{b=1}$ denotes a batch of $B^s$ samples randomly selected from $D^s_l$. $\mathcal{A}_w(\cdot)$  represents the weak augmentation applied to the input, and $H(\cdot, \cdot)$ is the cross-entropy function. $p_{w^s}(\cdot)$ denotes the softmax predicted probability distribution (confidence) produced by the server's model with parameters $w^s$ for the input data. Based on this loss, the server model's parameters are updated as follows at communication round $r$:

\begin{equation}
    w^{s}_{r} = w^{s}_{r} - \eta\,\nabla L^{s}_{l}.
\end{equation}

\subsection{Client} \label{subsec:client}
On the client side, there is a relatively larger amount of unlabeled data compared to the server, denoted by $\mathcal{D}^m_u = \{x^m_n\}^{D^m_u}_{n=1}$ which is used for training. According to the "generate pseudo-labels with global model" approach of SemiFL \cite{diao2022semifl}, the client receives a global model from the server and, using it, generates a pseudo-labeled dataset with a single pseudo-labeling step per training round. The client’s training loss at communication round $r$ is defined as follows, and the client updates its model parameters based on this loss:

\begin{equation} \label{eq:pl_dataset}
    \mathcal{D}^m_{p} = \left\{ (x^m_n, \hat{q}^g_n) \ \big| \ \max(q^g_{x^m_n}) > \tau\right\}^{D^m_u}_{n=1},
\end{equation}

\begin{equation} \label{eq:pl_loss}
    L^m_p = \frac{1}{B^m_{p}}\sum\limits_{(x^m_b, \hat{q}^g_b) \in \mathcal{B}^m_p}H(\hat{q}^g_{x^m_b}, p_{w^m}(y \mid \mathcal{A}_s(x^m_b))),
\end{equation}

In this equation, $\mathcal{B}^m_p = \{(x^m_b, \hat{q}^g_b)\}^{B^m_p}_{b=1}$ denotes a batch of size $B^m_p$ randomly sampled from the pseudo-labeled dataset $\mathcal{D}^m_p$ generated by the global model. Here, $q^g_{x^m_n}=p_{w^g}(y|A_w(x^m_n))$ is the notation for the softmax prediction probability distribution output by the global model with parameters $w^g$ for the weakly augmented input data $x^m_n$. $\max(\cdot)$ denotes the highest probability among the model’s output probabilities, and $\hat{q}$ represents the index corresponding to the highest predicted probability, where $\hat{q} = arg \max(p_w(\cdot))$. $\mathcal{A}_s(\cdot)$ denotes a strong augmentation, such as RandAugment \cite{cubuk2020randaugment} or CTAugment \cite{berthelot2019remixmatch}. Lastly, $\tau$ is a hyperparameter that determines whether a pseudo-label should be used when the prediction probability exceeds this threshold. Based on this loss, the server model's parameters are updated as follows at communication round $r$:

\begin{equation}
    w^m_r = w^m_r - \eta \nabla L^m_p.
\end{equation}

\begin{figure*}[!t]
\vspace{-5pt}
  \centering
  \includegraphics[width=\textwidth]{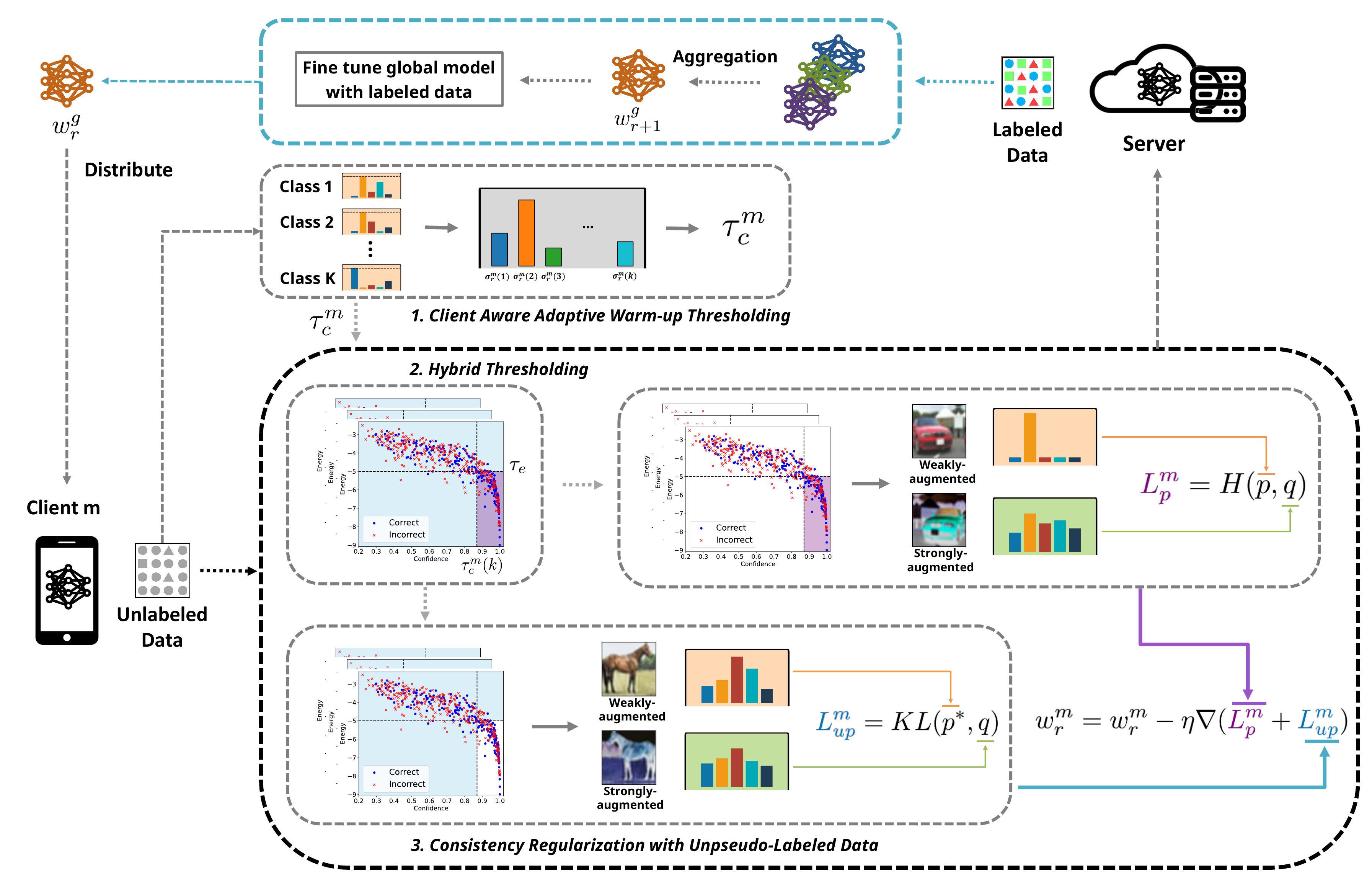}
  \caption{An overall pipeline of CATCHFed. \textcolor{cyan}{\textbf{Blue}}-bordered components are server-side processes, while \textcolor{gray}{\textbf{gray}} and \textcolor{black}{\textbf{black}}-bordered components are client-side processes.} 
  \label{fig:main_figure}
  \vspace{-5pt}
\end{figure*}

\subsection{Mixup loss} \label{sec:mixup}

In SemiFL \cite{diao2022semifl}, an additional Mixup loss is applied to enhance the generalization performance of the model. The composition and application of the Mixup loss are described in detail as follows.
We construct a pseudo-labeled dataset $\mathcal{D}^m_p$, as previously described in Eq. \ref{eq:pl_dataset}. Then, following the method of MixMatch \cite{berthelot2019mixmatch}, we construct the Mixup dataset via sampling with replacement as follows:

\begin{equation}
    \mathcal{D}^m_{mix} = \text{\textbf{SampleWithReplacement}}(\mathcal{D}^m_{p}, |\mathcal{D}^m_p|),
\end{equation}

In this process, the sampled dataset $\mathcal{D}^m_{mix}$ has the same size as the original dataset $\mathcal{D}^m_p$, thus satisfying $|\mathcal{D}^m_p|=|\mathcal{D}^m_{mix}|$.
Subsequently, each local client $m$ splits the datasets $\mathcal{D}^m_p$ and $\mathcal{D}^m_{mix}$ into batches $\mathcal{B}^m_p$ and $\mathcal{B}^m_{mix}$, respectively, for local training. Using the Mixup hyperparameter that determines the mixing ratio between the two datasets, the mixing ratio is sampled from a Beta distribution. Then the Mixup data is constructed through the following procedure:

\begin{equation} \label{eq:mixup_dataset}
    \lambda_{mix} \sim Beta(a, a), \quad x_{mix} \leftarrow \lambda_{mix} x_{p, b} \ + \ (1-\lambda_{mix})x_{mix, b} ,
\end{equation}

Here, $\alpha$ is the Mixup hyperparameter controlling the data mixing ratio, and following \cite{zhang2017mixup}, it is also set to 0.75 in this paper.
The loss for the constructed Mixup dataset is defined as follows:

\begin{equation} \label{eq:mixup_loss}
    \begin{aligned}
    L^m_{mix} 
    &= \frac{1}{B^m_{mix}} 
    \sum_{(x^m_{mix,b},\hat{q}^g_{p,b},\hat{q}^g_{mix,b}) \in \mathcal{B}^m_{mix}} \\[4pt]
    &\quad \lambda_{mix} \cdot 
    H\!\left(\hat{q}^g_{p,b},\ 
    p_{w^m}\!\left(y \mid \mathcal{A}_w(x^m_{mix,b})\right)\right) \\[4pt]
    &\quad + (1 - \lambda_{mix}) \cdot 
    H\!\left(\hat{q}^g_{mix,b},\ 
    p_{w^m}\!\left(y \mid \mathcal{A}_w(x^m_{mix,b})\right)\right),
    \end{aligned}
\end{equation}

A key characteristic of the Mixup dataset is that, unlike the pseudo-labeled data in Eq. \ref{eq:pl_loss}, it applies weak augmentation. Finally, the resulting Mixup loss $L_{mix}$ is combined with the existing losses $L_p$ to define the final unsupervised learning loss $L_u$ as follows:

\begin{equation}
    L^m_u = L^m_p + L^m_{mix}.
\end{equation}

\section{Proposed methods}
In this section, we propose \textbf{CATCHFed}, designed to effectively utilize clients' unlabeled data. CATCHFed consists of three main components: 1) \textit{\textbf{C}lient-Aware \textbf{A}daptive Warm-up \textbf{T}hresholding}, which mitigates early-stage confirmation bias and dynamically adjusts thresholds considering class-wise learning difficulty at each client to make effective use of unlabeled data, 2) \textit{\textbf{C}onsistency Regularization with Unpseudo-labeled Data}, which effectively leverages unlabeled data not selected as pseudo-labels through consistency regularization, and 3) \textit{\textbf{H}ybrid Thresholding}, which introduces an additional energy score-based threshold to enhance pseudo-label quality. The overall pipeline is illustrated in Fig. \ref{fig:main_figure}.

\begin{figure*}[t!]
\vspace{-5pt}
  \centering
  \begin{subfigure}[b]{0.32\linewidth}
    \centering
    \includegraphics[width=\linewidth]{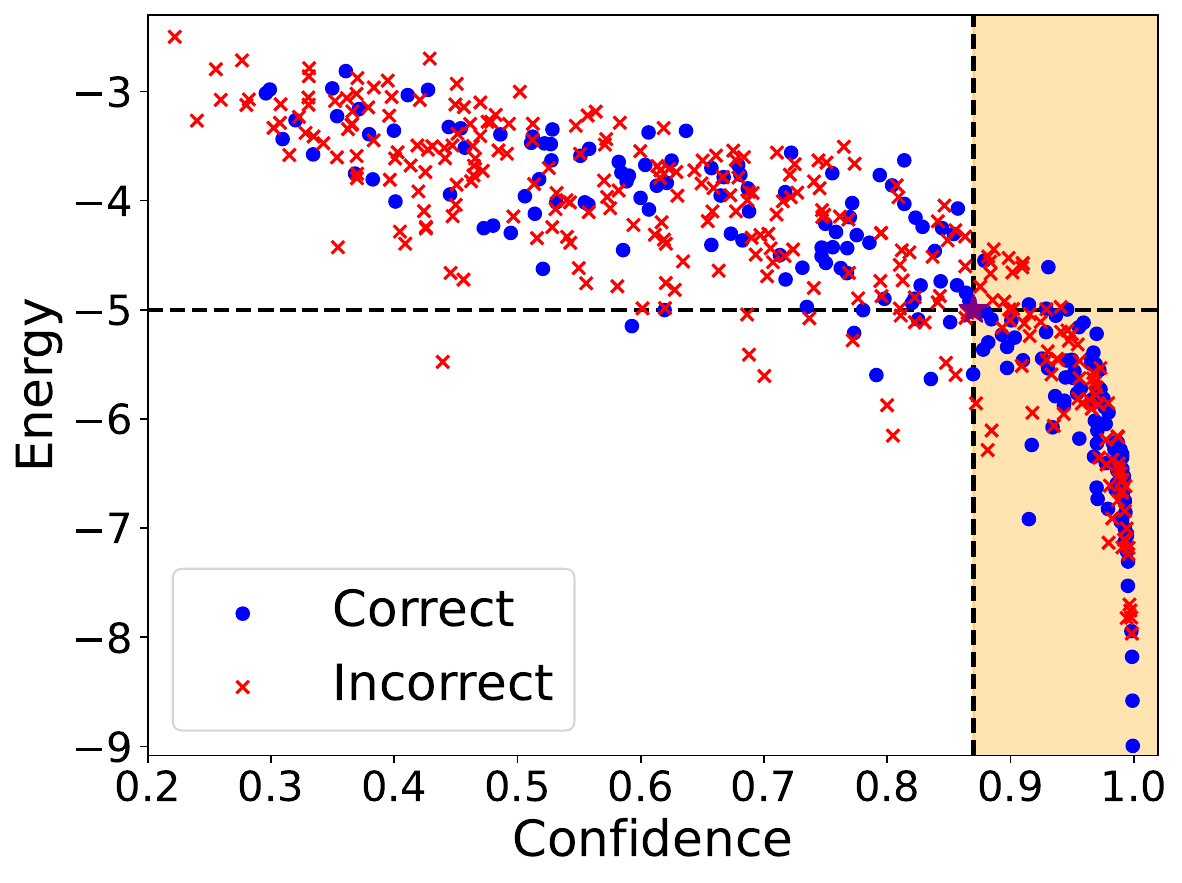}
    \caption{Confidence-based thresholding}
    \label{fig:conf_based_thresholding}
  \end{subfigure}
  \hfill
  \begin{subfigure}[b]{0.32\linewidth}
    \centering
    \includegraphics[width=\linewidth]{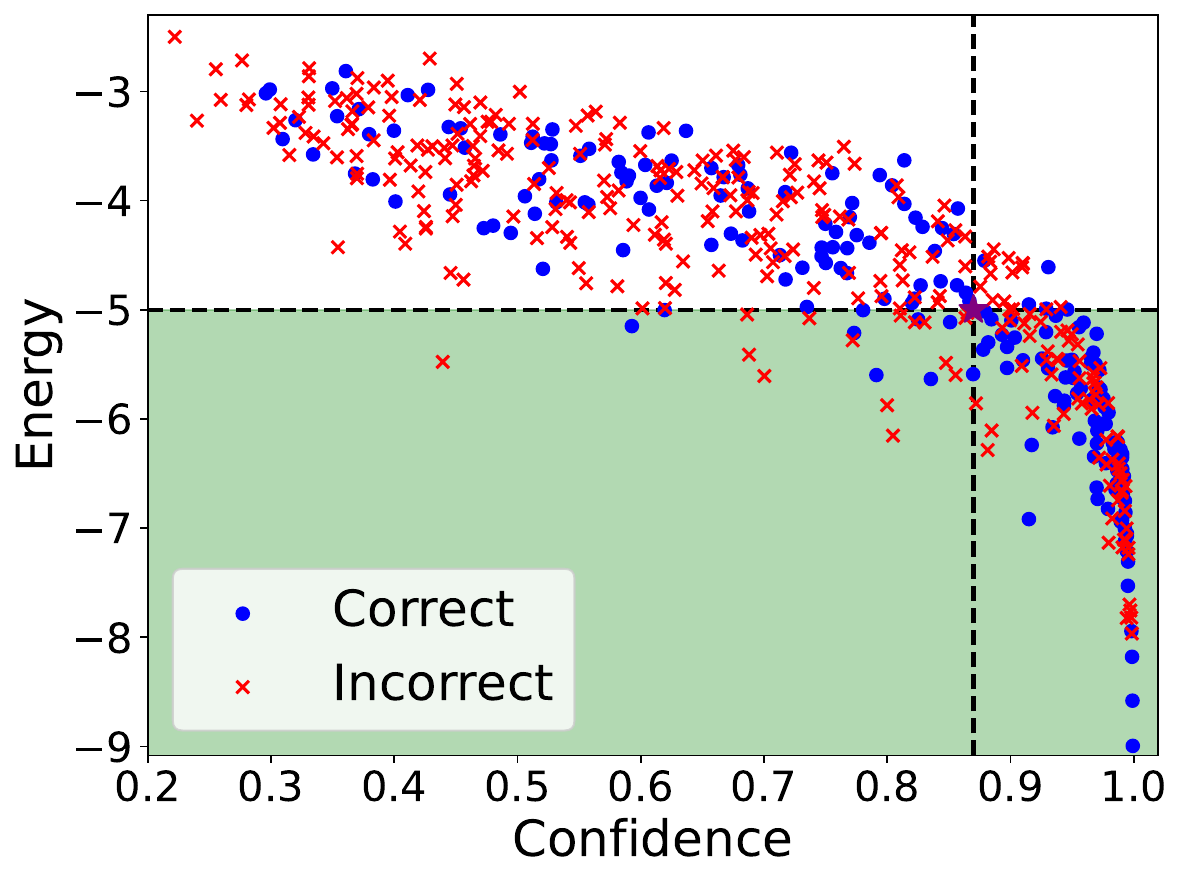}
    \caption{Energy-based thresholding}
    \label{fig:energy_based_thresholding}
  \end{subfigure}
  \hfill
  \begin{subfigure}[b]{0.32\linewidth}
    \centering
    \includegraphics[width=\linewidth]{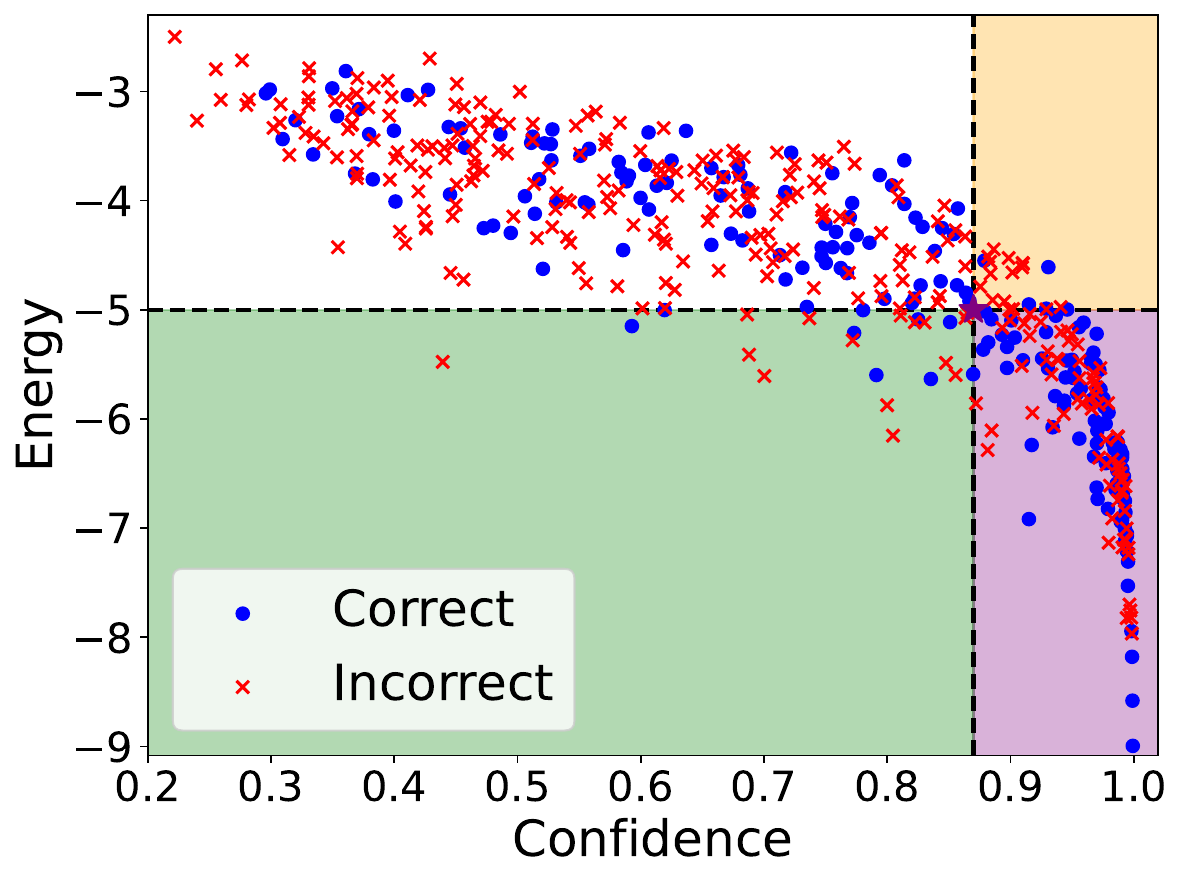}
    \caption{Hybrid thresholding}
    \label{fig:hybrid_thresholding}
  \end{subfigure}
  \caption{Visualization of pseudo-label selection regions by (a) confidence-based, (b) energy-based, and (c) hybrid thresholding. Shaded regions indicate unlabeled samples selected for pseudo-labeling.}
  \label{fig:visualization_hybrid_thresholding}
  \vspace{-5pt}
\end{figure*}

\subsection{Client-aware adaptive warm-up thresholding}
Existing class-wise fixed-threshold-based methods use unlabeled data when the prediction confidence exceeds a predefined threshold. 
This works well for SSL, but for decentralized clients, this global threshold may cause significant drop of pseudo-labels for some clients because the threshold does not consider differences between clients' data.
Inspired by FlexMatch \cite{zhang2021flexmatch}, we propose Client‑Aware Adaptive Warm‑up Thresholding (CAWT), which adjusts per‑class thresholds per client based on dataset difficulty and includes more samples into pseudo-labels in the warm‑up phase. Each client computes class‑wise difficulty:

\begin{equation} \label{eq:get_sigma}
    \sigma^m_r(k) = \sum\limits_{x^m_u \in \mathcal{D}^m_u} \mathbbm{1}(\max(q^g_{x^m_u}) > \tau) \cdot \mathbbm{1}(\hat{q}^g_{x^m_u} = k),
\end{equation}

where $r$ is the communication round and $\mathbbm{1}(\cdot)$ is the indicator function. $\sigma^m_r(k)$ measures how confidently the model predicts class $k$, serving as a proxy for that class’s learning difficulty. To normalize $\sigma^m_r(k)$ to $[0, 1]$, we define:

\begin{equation} \label{eq:beta}
    \beta^m_r(k) = 
    \begin{cases}
    \dfrac{\sigma^m_r(k)}{\sigma^m_{\text{rest}, r}}, & \text{if } \sum\limits_{k} \sigma^m_r < \sigma^m_{\text{rest}, r} \\
    \\[-1ex]
    \dfrac{\sigma^m_r(k)}{\max\limits_{k} \sigma^m_r}, & \text{otherwise}
    \end{cases},
\end{equation}

\begin{equation} \label{eq:at_tau}
    \tau^m_c(k) = \pi(\beta^m_r(k)) \cdot \tau.
\end{equation}

Here, we let $\sigma^m_{\text{rest}, r} = D^m_u - \sum_k\sigma^m_r(k)$, which counts how many unlabeled samples on client $m$ at round $r$ fail to exceed the global threshold $\tau$. When $\sigma^m_{rest, r}$ dominates during normalization, it is used as the denominator to lower the per-class threshold $\tau^m_c(k)$, allowing more unlabeled data to be utilized in the early stages of training. This \textit{warm‑up phase} speeds up initial learning and helps mitigate confirmation bias. To set the final confidence‑based adaptive threshold, we apply a nonlinear mapping function $\pi(\cdot)$ allowing $\tau^m_c(k)$ to increase smoothly rather than abruptly. Although various mapping functions could be used, in this work we use $\pi(x) = \frac{x}{2-x}$ as in the FlexMatch \cite{zhang2021flexmatch}.

\subsection{Hybrid thresholding}
In training methods using pseudo-label, the quality (accuracy) of pseudo-label is one of key components. As shown in Fig. \ref{fig:ssl_vs_ssfl_pl_acc}, pseudo-label quality of conventional SSFL is much lower than standard SSL methods under scarce labeled data. To address this, we extend the idea from \cite{yu2023inpl} by introducing an energy‑score‑based threshold alongside the existing confidence‑based adaptive threshold, yielding a hybrid thresholding scheme. As described in Section \ref{subsec:energy_score}, the energy score measures how close an input sample is to the current training distribution composed of labeled and pseudo‑labeled data. Lower energy values indicate greater conformity to the learned data distribution. Under hybrid thresholding, a pseudo-label is generated only when both conditions are satisfied:

\begin{align} \label{eq:pseudo_labeld_dataset}
    \mathcal{D}^m_p = \big\{ (x^m_n, \hat{q}^g_n) \ \big| \ & \max(q^g_n) > \tau^m_c(\hat{q}^g_n) \notag \\
    & \land \ E(f^{w^g}(\mathcal{A}_w(x^m_n))) < \tau_e \big\}^{D^m_u}_{n=1},
\end{align}

Here, $E(\cdot)$ denotes the energy score function, and $\tau_e$ is a fixed energy threshold hyperparameter. In other words, a sample is pseudo‑labeled only if the model is sufficiently confident and the sample lies close to the training distribution, ensuring higher pseudo‑label quality. The resulting data distribution is illustrated in Fig.\ref{fig:visualization_hybrid_thresholding}. The pseudo‑label loss is calculated as follows:

\begin{equation} \label{eq:hpl_loss}
    L^m_{p} = \frac{1}{B^m_p}\sum\limits_{(x^m_{b}, \hat{q}^g_b) \in \mathcal{B}^m_p}H(\hat{q}^g_b, p_{w^m}(y \mid \mathcal{A}_s(x^m_b))),
\end{equation}

\begin{algorithm}[!t]
\DontPrintSemicolon
\SetKwProg{ServerUpdate}{Procedure \textnormal{ServerUpdate}}{}{}
\SetKwProg{ClientUpdate}{Procedure \textnormal{ClientUpdate}}{}{}

\KwIn{total training rounds $R$, local epoch $E$, train iteration $I$ labeled dataset $\mathcal{D}^s_l$, unlabeled dataset $\mathcal{D}^{m:M}_u$, total number of clients $M$, activation rate $P$, confidence threshold $\tau$, energy threshold $\tau_e$, learning rate $\eta$, server batch size $B^s$, client batch size $B^m_p$, ratio of pseudo-labeled to unpseudo-labeled data $\mu$}
\KwOut{Final global model $w^g_R$}

Initialize variables \;
\For{$r = 1$ \KwTo $R$}{
    $w^s_r \leftarrow$ \textnormal{\textbf{ServerUpdate}}($\mathcal{D}^s_l, w^g_{r}$)\;
    Update \textnormal{sBN} statistics\;
    $S_r \leftarrow$ Randomly sample $M\!\cdot\!P$ clients\;
    \ForPar{each client $m \in S_r$}{
        $w^m_r \leftarrow$ \textnormal{\textbf{ClientUpdate}}($\mathcal{D}^m_u, w^s_r$)\;
    }
    $w^g_{\,r+1} \leftarrow \sum_{m=1}^{M} a^m_r\,w^m_r$\ \tcp{$a^m :$ aggregation weight} \;
}
\Return $w^g_R$\;

\BlankLine
\ServerUpdate{$(\mathcal{D}^s_l,w^g_r)$}{
    \For{$e = 1$ \KwTo $E$}{
        Divide $\mathcal{D}^s_l$ into batches $\{ \mathcal{B}^s \}$ of size $B^s$ \;
        \For{$i = 1$ \KwTo $|\{\mathcal{B}^s\}|$}{
            use Eq. \ref{eq:l_loss} to compute $L^s_l$\;
            $w^s_r \leftarrow w^s_r - \eta\,\nabla L^s_l$\;
        }
    }
    \KwRet $w^s_r$\;
}

\BlankLine
\ClientUpdate{$(\mathcal{D}^m_u, w^s_r)$}{
    use Eq. \ref{eq:get_sigma}, \ref{eq:beta}, \ref{eq:at_tau} to compute $\tau^m_c(k)$ \;
    \If{$\sum_k\sigma^m_r < \sigma^m_{rest, r}$ \tcp{warm-up phase}}{
        $\mathcal{D}^m_p = \left\{ (x^m_b, \hat{q}^g_b) \ \big| \max(q^g_b) > \tau^m_c(\hat{q}^g_b)\right\}^{D^m_u}_{b=1}$ \;
    }
    \Else{
        use Eq. \ref{eq:pseudo_labeld_dataset}, \ref{eq:unpseudo_labeled_dataset} to get $\mathcal{D}^m_p$, $\mathcal{D}^m_{up}$ \;
    }
    $\mathcal{D}^m_{mix} = \text{\textbf{SampleWithReplacement}}(\mathcal{D}^m_{p}, |\mathcal{D}^m_p|)$ \;
    \For{$i = 1$ \KwTo $I$}{
        $\mathcal{B}^m_p \leftarrow$ Randomly sample $B^m_p$ from $\mathcal{D}^m_p$\;
        $\mathcal{B}^m_{up} \leftarrow$ Randomly sample $\mu B^m_p$ from $\mathcal{D}^m_{up}$\;
        $\mathcal{B}^m_{mix} \leftarrow$ Randomly sample $B^m_p$ from $\mathcal{D}^m_{mix}$\;
        Generate the Mixup dataset using Eq. \ref{eq:mixup_dataset}. \;
        use Eq. \ref{eq:hpl_loss}, \ref{eq:upl_loss}, \ref{eq:mixup_loss} to compute $L^m_p$, $L^m_{up}$, $L^m_{mix}$\;
        $L^m_u = L^m_p + L^m_{up} + L^m_{mix}$ \;
        $w^m_r = w^m_r - \eta \nabla L^m_u$ \;
    }
    \KwRet $w^m_r$\;
}
\caption{CATCHFed} \label{alg:catchfed}
\end{algorithm}

\subsection{Consistency regularization with unpseudo-labeled data}

As shown in Fig. \ref{fig:ssl_vs_ssfl_util_ratio}, unlabeled data utilization in SSFL environments is significantly lower than in standard SSL methods. SequenceMatch \cite{nguyen2024sequencematch} shows that even uncertain samples (those that fail to meet the confidence threshold) can achieve competitive performance, indicating that low-confidence samples can still serve as valuable learning resources. Based on this observation, we aim to make the most of discarded unlabeled data in SSFL by incorporating unpseudo-labeled samples that do not satisfy the hybrid threshold into consistency regularization, using their predictions as soft labels.

\begin{align} \label{eq:unpseudo_labeled_dataset}
    \mathcal{D}^m_{up} = \big\{ (x^m_n, q^g_n) \ \big| \ & \max(q^g_n) \leq \tau^m_c(\hat{q}^g_n) \notag \\
    & \lor \ E(f^{w^g}(\mathcal{A}_w(x^m_n))) \geq \tau_e \big\}^{D^m_u}_{n=1},
\end{align}

Here, $\mathcal{D}^m_{up}$ denotes the set of all unlabeled samples that fail the Hybrid threshold. We then compute the consistency‑regularization loss over a randomly sampled Unpseudo-labeled batch:

\begin{equation} \label{eq:upl_loss}
    L^m_{up} = \frac{1}{B^m_{up}}\sum\limits_{(x^m_{b}, q^g_b) \in \mathcal{B}^m_{up}}KL(q^g_b, p_{w^m}(y \mid \mathcal{A}_s(x^m_b))),
\end{equation}

$\mathcal{B}^m_{up} = \{(x^m_b, q^g_b)\}^{\mu B^m_p}_{b=1}$ represents a batch randomly sampled from $\mathcal{D}^m_{up}$, where $\mu$ denotes the ratio between the $B^m_p$ and the $B^m_{up}$ within the total batchsize $B^m_p + B^m_{up}$. $KL(\cdot, \cdot)$ denotes the KL divergence loss, and by using soft labels instead of hard labels for weakly augmented inputs, the model is encouraged to align its predictions on strongly augmented inputs with these soft targets. This helps the model produce consistent predictions across various input perturbations, thereby improving generalization performance, while also mitigating confirmation bias that may arise from training repeatedly on only easy, high-confidence samples. Finally, the overall unlabeled loss is defined as follows:

\begin{equation}
    L^m_u = L^m_{p} + L^m_{up} + L^m_{mix}.
\end{equation}

During warm-up phase, hybrid thresholding is disabled so that the unlabeled data usage is maximized. The overall algorithm is detailed in Algorithm \ref{alg:catchfed}.

\begin{table}[t]
\vspace{-10pt}
  \caption{Experimental settings}
  \label{tab:experiment_settings}
  \centering
  \resizebox{\linewidth}{!}{
  \begin{tabular}{c c c c c c c c c}
    \toprule
    \multicolumn{3}{c}{\textbf{Dataset}} & \multicolumn{2}{c}{\textbf{CIFAR10}} & \multicolumn{2}{c}{\textbf{SVHN}} & \multicolumn{2}{c}{\textbf{CIFAR-100}} \\
    \midrule
    \multicolumn{3}{c}{Number of labeled data} & 20 & 40 & 20 & 40 & 200 & 400 \\
    \midrule
    \multicolumn{3}{c}{Model} & \multicolumn{4}{c}{WResNet28x2} & \multicolumn{2}{c}{WResNet28x8} \\
    \midrule
    \multicolumn{3}{c}{Number of total clients } & \multicolumn{6}{c}{100} \\
    \midrule
    \multicolumn{3}{c}{Participation ratio } & \multicolumn{6}{c}{0.1} \\
    \midrule
    \multicolumn{3}{c}{Communication rounds } & \multicolumn{6}{c}{800} \\
    \midrule
    \multirow{3}{*}{\textbf{Server}} 
        & \multirow{2}{*}{Others} & Batchsize & \multicolumn{6}{c}{10} \\
        \cmidrule(r){3-9}
        & & Epoch             & \multicolumn{6}{c}{5} \\
      \cmidrule(r){2-9}
        & Ours & Train Iter   & \multicolumn{6}{c}{50} \\
    \midrule
    \multirow{11}{*}{\textbf{Client}} 
        & \multirow{2}{*}{Others} & Batchsize & \multicolumn{6}{c}{10} \\
      \cmidrule(r){3-9}
        & & Epoch         & \multicolumn{6}{c}{5} \\
        \cmidrule(r){2-9}
        & \multirow{2}{*}{$(FL)^2$} & Batchsize & \multicolumn{6}{c}{32} \\
      \cmidrule(r){3-9}
        & & Epoch         & \multicolumn{6}{c}{5} \\
      \cmidrule(r){2-9}
        & \multirow{4}{*}{Ours} & Warm-up iter & 100 & 50 & 100 & 50 & 100 & 50 \\
      \cmidrule(r){3-9}
        & & Train iter & \multicolumn{6}{c}{100} \\
      \cmidrule(r){3-9}
        & & $\mu$ & \multicolumn{6}{c}{1} \\
      \cmidrule(r){3-9}
        & & Energy threshold & \multicolumn{2}{c}{-5.0} & \multicolumn{2}{c}{-7.0} & \multicolumn{2}{c}{-6.5} \\
      \cmidrule(r){2-9}
        & \multicolumn{2}{c}{Confidence threshold} & \multicolumn{6}{c}{0.95} \\
      \midrule
      \multirow{5}{*}{\shortstack{\textbf{Server} \\ \& \\ \textbf{Client}}}  & \multicolumn{2}{c}{Optimizer}         & \multicolumn{6}{c}{SGD} \\
      \cmidrule(r){2-9}
        & \multicolumn{2}{c}{Learning rate}     & \multicolumn{6}{c}{3.0E-02} \\
      \cmidrule(r){2-9}
      & \multicolumn{2}{c}{Weight decay}      & \multicolumn{6}{c}{5.0E-04} \\
      \cmidrule(r){2-9}
        & \multicolumn{2}{c}{Momentum}          & \multicolumn{6}{c}{0.9} \\
    \midrule
    \multirow{2}{*}{\textbf{Global}} & \multicolumn{2}{c}{Momentum}          & \multicolumn{6}{c}{0.5} \\
             \cmidrule(r){2-9}
       & \multicolumn{2}{c}{Scheduler}          & \multicolumn{6}{c}{Cosine Annealing} \\
    \bottomrule
  \end{tabular}
  }
  \vspace{-10pt}
\end{table}

\begin{table*}[t]
    \centering
    \caption{Test accuracy on CIFAR-10/100 and SVHN under IID and Non-IID settings (averaged over three random seeds). \textbf{Bold} indicates the highest accuracy, \underline{underline} indicates the second-highest.}
    \begin{tabular}{c c c c c c c c}
        \toprule
        \multicolumn{2}{c}{\textbf{Dataset}} & \multicolumn{2}{c}{\textbf{CIFAR-10}} & \multicolumn{2}{c}{\textbf{CIFAR-100}} & \multicolumn{2}{c}{\textbf{SVHN}} \\
        \midrule
        \multicolumn{2}{c}{\textbf{Number of labeled data}} & \textbf{20} & \textbf{40} & \textbf{200} & \textbf{400} & \textbf{20} & \textbf{40} \\
        \midrule
        \multirow{4}{*}{\textbf{Non-IID}} & \textbf{FedMatch} & $20.24 (3.24)$ & $25.98 (0.97)$ & $9.4 (0.69)$ & $13.17 (0.45)$ & $13.13 (2.13)$ & $13.34 (1.98)$ \\
        & \textbf{FedCon} & $\underline{24.7} (3.01)$ & $32.82 (0.72)$ & $18.72 (1.21)$ & $25.12 (0.65)$ & $\textbf{20.6} (5.18)$ & $\underline{41.06} (3.68)$ \\
        & \textbf{SemiFL} & ${23.72} (1.84)$ & $\underline{35.65} (4.19)$ & $17.86 (0.36)$ & $25.84 (0.61)$ & $16.66 (3.33)$ & $22.06 (2.66)$ \\
        \textbf{$(Dir(0.1))$} & \textbf{$(FL)^2$} & $21.66 (0.73)$ & $33.47 (1.7)$ &  $\underline{21.9} (2.07)$ & $\underline{26.62} (0.51)$ & $16.65 (1.93)$ & $22.99 (1.41)$ \\
        & \cellcolor{LimeGreen!50}\textbf{CATCHFed} & \cellcolor{LimeGreen!50}$\textbf{29.55}(4.28)$ & \cellcolor{LimeGreen!50}$\textbf{40.74}(3.05)$ & \cellcolor{LimeGreen!50}$\textbf{24.59} (0.67)$ & \cellcolor{LimeGreen!50}$\textbf{32.86} (0.96)$ & \cellcolor{LimeGreen!50}$\underline{20.56} (4.47)$ & \cellcolor{LimeGreen!50}$\textbf{42.67} (2.77)$ \\
        \midrule
        \multirow{4}{*}{\textbf{Non-IID}} & \textbf{FedMatch} & $19.72 (2.64)$ & $26.1 (1.41)$ & $9.47 (0.56)$ & $12.99 (0.27)$ & $11.25 (1.9)$ & $12.5 (1.0)$ \\
        & \textbf{FedCon} & $24.19 (0.97)$ & $35.1 (0.59)$ & $21.47 (0.96)$ & $\underline{29.56} (0.76)$ & $\underline{23.41} (5.8)$ & $\underline{45.25} (5.39)$ \\
        & \textbf{SemiFL} & $24.28 (4.15)$ & $41.83 (5.14)$ & $19.36 (0.33)$ & $28.16 (0.24)$ & $17.55 (4.15)$ & $37.92 (9.74)$ \\
        \textbf{$(Dir(0.3))$} & \textbf{$(FL)^2$} & $\underline{29.25} (5.51)$ & $\underline{44.68} (5.6)$ &  $\underline{25.51} (1.01)$ & $29.24 (1.05)$ & $15.41 (1.59)$ & $31.07 (4.05)$ \\
        & \cellcolor{LimeGreen!50}\textbf{CATCHFed} & \cellcolor{LimeGreen!50}$\textbf{31.78}(8.39)$ & \cellcolor{LimeGreen!50}$\textbf{46.42}(3.68)$ & \cellcolor{LimeGreen!50}$\textbf{30.28} (0.41)$ & \cellcolor{LimeGreen!50}$\textbf{39.59} (1.5)$ & \cellcolor{LimeGreen!50}$\textbf{24.13} (9.83)$ & \cellcolor{LimeGreen!50}$\textbf{54.6} (3.91)$ \\
        \midrule
        \multirow{5}{*}{\textbf{IID}} & \textbf{FedMatch} & $18.75 (2.54)$ & $24.99 (2.14)$ & $9.73 (0.69)$ & $12.5 (0.86)$ & $11.89 (1.61)$ & $14.07 (2.2)$ \\
        & \textbf{FedCon} & $26.46 (3.1)$ & $39.33 (1.14)$ & $25.27 (0.82)$ & $\underline{32.53} (0.45)$ & $27.19 (8.28)$ & $59.61 (3.81)$ \\
        & \textbf{SemiFL} & $35.11 (7.44)$ & $65.19 (1.55)$ & $20.16 (0.56)$ & $31.04 (0.55)$ & $\underline{31.85 (13.63)}$ & $\underline{66.2} (11.83)$ \\
        & \textbf{$(FL)^2$} & $\underline{41.12} (2.84)$ & $\underline{75.2} (1.97)$ & $\underline{26.95} (2.89)$ & $31.62 (1.37)$ & $20.75 (5.89)$ & $58.62 (13.77)$ \\
        & \cellcolor{LimeGreen!50}\textbf{CATCHFed} & \cellcolor{LimeGreen!50}$\textbf{46.33}(5.33)$ & \cellcolor{LimeGreen!50}$\textbf{77.45}(9.4)$ & \cellcolor{LimeGreen!50}$\textbf{36.75} (1.21)$ & \cellcolor{LimeGreen!50}$\textbf{47.09} (0.08)$ & \cellcolor{LimeGreen!50}$\textbf{34.11} (16.9)$ & \cellcolor{LimeGreen!50}$\textbf{85.61} (4.4)$ \\
        \bottomrule
    \end{tabular}
    \label{tab:main_table}
\end{table*}

\section{Experimental results} \label{sec:exp_results}

\subsection{Experimental setup} \label{sec:exp_setup}

To evaluate the proposed method, we conduct experiments on three datasets including CIFAR-10, CIFAR-100 \cite{krizhevsky2009learning}, and SVHN \cite{netzer2011reading}. We simulate an extremely low-label setting by using 20 and 40 labeled samples for CIFAR-10 and SVHN, and 200 and 400 for CIFAR-100. For the Non-IID setting, client data is sampled using a Dirichlet distribution \cite{hsu2019measuring} to reflect data imbalance. We simulate 800 communication rounds with 100 clients, where the client participation rate is set to $P = 0.1$. Other settings follow the configuration of SemiFL \cite{diao2022semifl}. All SSFL methods are implemented using the Flower framework \cite{beutel2020flower} for fair comparison, and SSL methods are evaluated with the USB \cite{wang2022usb} benchmark suite. WideResNet-28x2 \cite{zagoruyko2016wide} is used as the backbone model for CIFAR-10 and SVHN, and WideResNet-28x8 is used for CIFAR-100. We use Nvidia RTX 4090 GPUs and Intel i7-13700F CPU for experiments. Most training hyperparameters follow the setup introduced
by SemiFL \cite{diao2022semifl}, with specifics outlined in Table \ref{tab:experiment_settings}. The optimizer and scheduler are SGD and cosine annealing, consistent with FixMatch \cite{sohn2020fixmatch} and SemiFL \cite{diao2022semifl}. Additionally, we utilize  momentum aggregation for global model updates, incorporating the gradient from the previous round with a momentum of factor of 0.5.

In our proposed method, the energy threshold $\tau_e$ is set to $-5.0$, $-7.0$, and $-6.5$ for CIFAR-10, SVHN, and CIFAR-100, respectively, while the ratio parameter $\mu$ is set to $1$. Further details on these hyperparameter configurations are provided in Sections \ref{sec:ablation_study} and \ref{sec:energy_threshold_setting}. To mitigate the increased confirmation bias caused by extremely limited labels, the warm-up phase is extended to 100 local training iterations in low-label settings. We also apply mixup loss for fair comparison with SemiFL and adopt the static batch normalization (sBN) from \cite{diao2020heterofl}. The detailed hyperparameter configuration is described in Table \ref{tab:experiment_settings}.

\begin{figure}[t]
\vspace{-10pt}
  \centering
  \subcaptionbox{Pseudo-label accuracy\label{fig:ht_macc}}
    [0.48\linewidth]{\includegraphics[width=\linewidth]{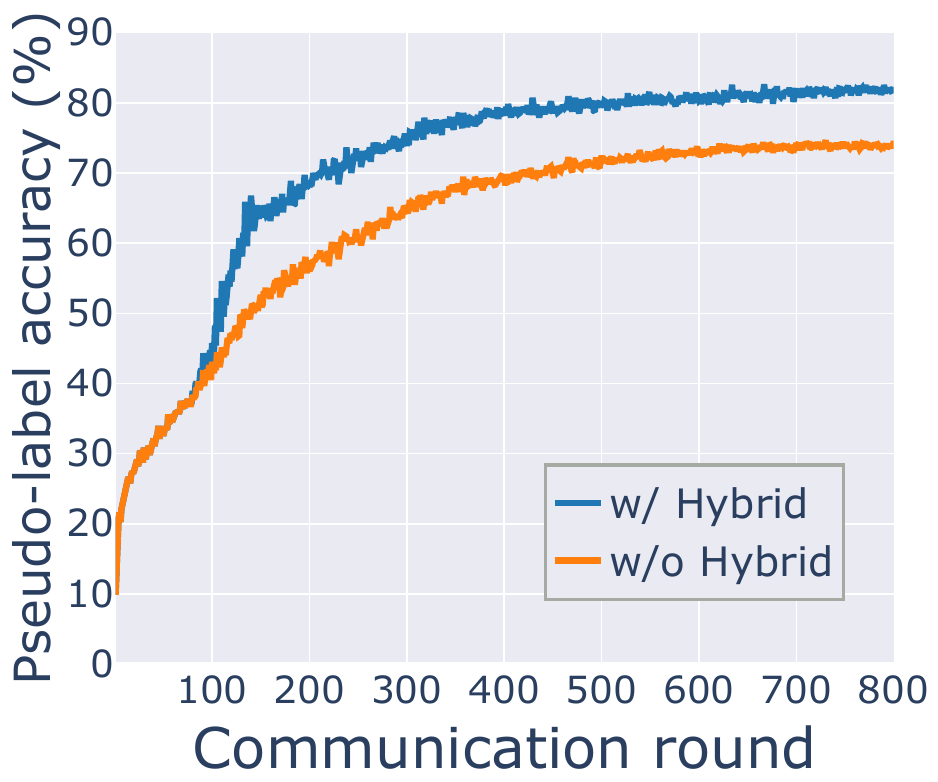}}
  \hfill
  \subcaptionbox{Wrong label ratio\label{fig:ht_wrong_ratio}}
    [0.48\linewidth]{\includegraphics[width=\linewidth]{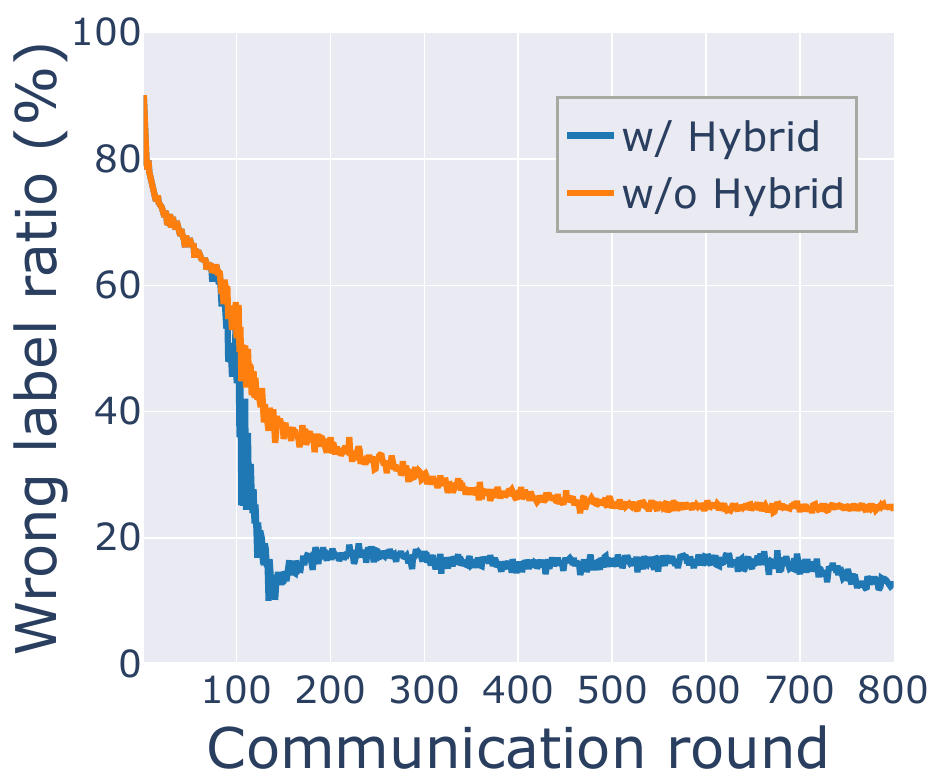}}
  \caption{Impact of Hybrid Thresholding on pseudo-label quality (CIFAR-10, IID 40 labels).}
  \label{fig:pl_quality_compare}
\vspace{-5pt}
\end{figure}

\begin{figure}[!t]
\vspace{-10pt}
  \centering
  \includegraphics[width=\columnwidth]{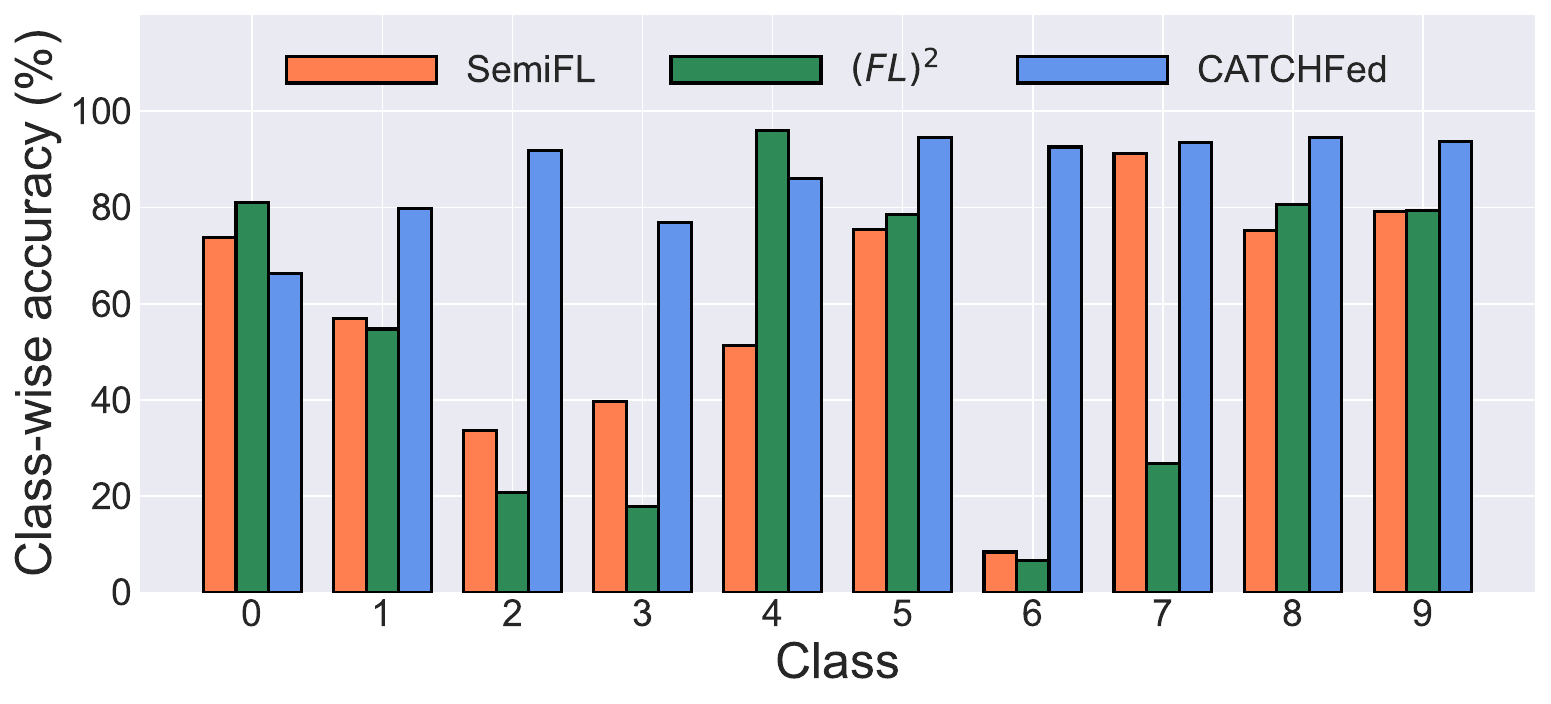}
  \caption{Comparison of classwise test accuracy for SemiFL, $(FL)^2$, and CATCHFed (SVHN, IID 40 labels).}
  \label{fig:classwise_acc_compare}
\vspace{-5pt}
\end{figure}

\subsection{Main results} \label{sec:main_results}

We compare the effectiveness of CATCHFed with representative SSFL methods: FedMatch \cite{jeong2020federated}, FedCon \cite{long2021fedcon}, SemiFL \cite{diao2022semifl}, and $(FL)^2$ \cite{lee20242}, and each method is implemented based on its official implementation. While the original FedCon uses two weak augmentations as input for client-side contrastive learning, in this study, we modify it based on FixMatch \cite{sohn2020fixmatch} to employ one weak and one strong augmentation. All experiments are carried out under identical settings using three different random seeds, and we present the average of the best accuracy to ensure fair comparison.

Table \ref{tab:main_table} summarizes results on CIFAR-10, CIFAR-100, and SVHN across varying labeled samples and data distributions (IID and Non-IID). CATCHFed consistently achieves superior performance over all baselines in most configurations. The improvement is particularly pronounced in IID scenarios. For example, on CIFAR-10, CATCHFed exceeds the best baseline by up to $5.21$ percentage points (pp). Furthermore, on CIFAR-10 with $Dir(0.1)$ and 20 labels, CATCHFed achieves $40.74\%$ accuracy, surpassing the best baseline by over $4.85$ pp. On SVHN with 40 labels, it improves the accuracy by $19.41$ pp, significantly surpassing the performance of existing methods. In addition, on CIFAR-100, it achieves up to $14.56$ pp improvement, indicating effectiveness even with many classes. These results indicate that CATCHFed effectively mitigates confirmation bias and utilizes unlabeled data efficiently, especially in extremely low-label cases. Overall, CATCHFed outperforms all baselines in both IID and Non-IID settings where a significant performance improvements was shown especially show in IID case.

\subsection{Effectiveness of hybrid thresholding}
Hybrid Thresholding is designed to strictly evaluate not only the model’s confidence but also whether the data samples belong to the current training distribution (in-distribution), aiming to enhance the quality of pseudo-labels. To verify the effectiveness of Hybrid Thresholding in improving pseudo-label quality, we compared pseudo-label accuracy as shown in Fig. \ref{fig:pl_quality_compare}. Fig. \ref{fig:ht_macc} illustrates that using Hybrid Thresholding significantly increases pseudo-label accuracy. Additionally, Fig. \ref{fig:ht_wrong_ratio} demonstrates a gradual decrease in the proportion of incorrect pseudo-labels, indicating that Hybrid Thresholding effectively filters out incorrect pseudo-labels as intended. These results confirm that Hybrid Thresholding effectively improves pseudo-label quality, leading to more stable training.

\begin{figure*}[t]
\vspace{-5pt}
  \centering
  \begin{subfigure}[b]{0.3\linewidth}
    \centering
    \includegraphics[width=\linewidth]{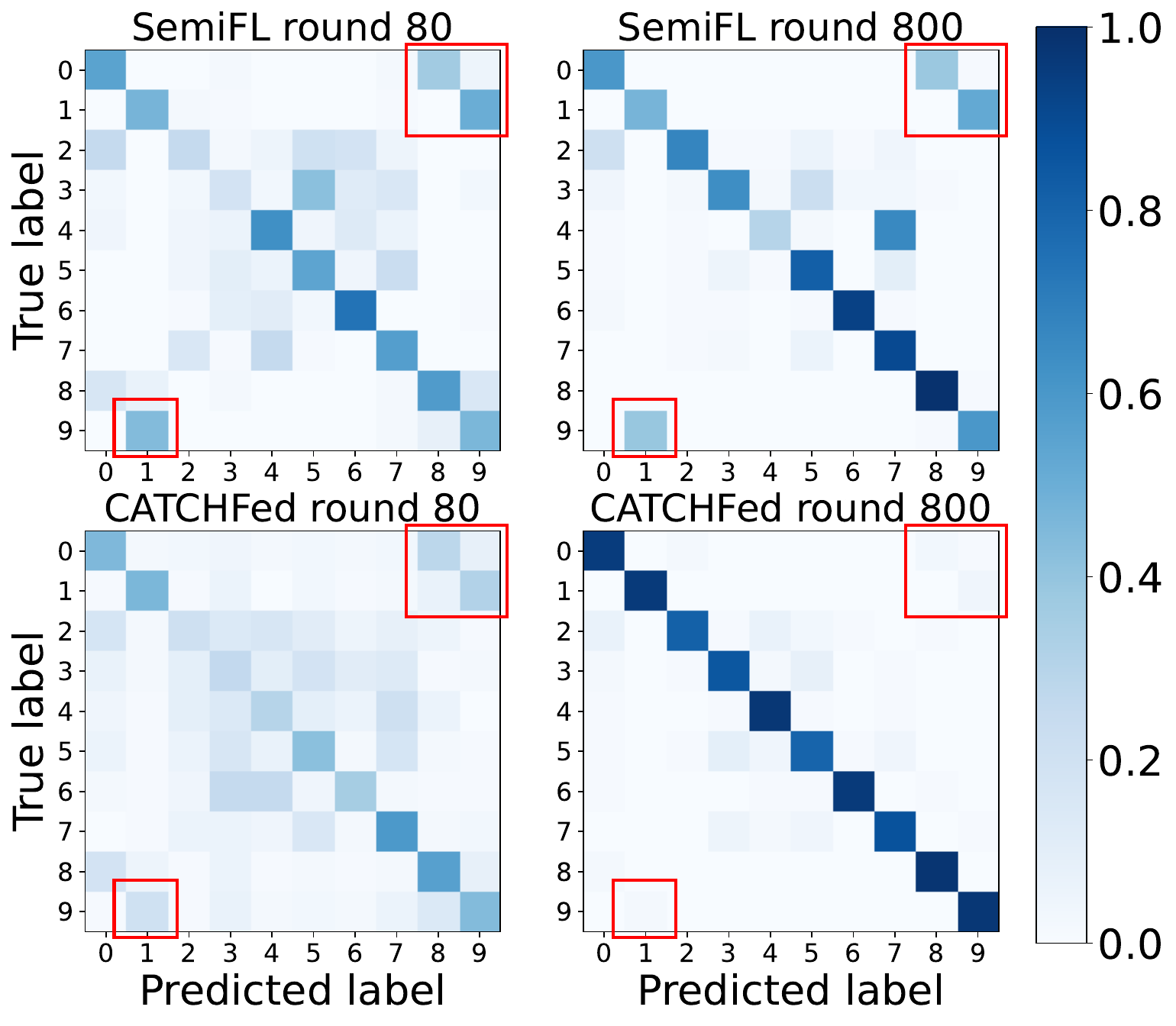}
    \caption{Pseudo-label confusion matrix}
    \label{fig:pl_conf_mat}
  \end{subfigure}
  \hfill
  \begin{subfigure}[b]{0.32\linewidth}
    \centering
    \includegraphics[width=\linewidth]{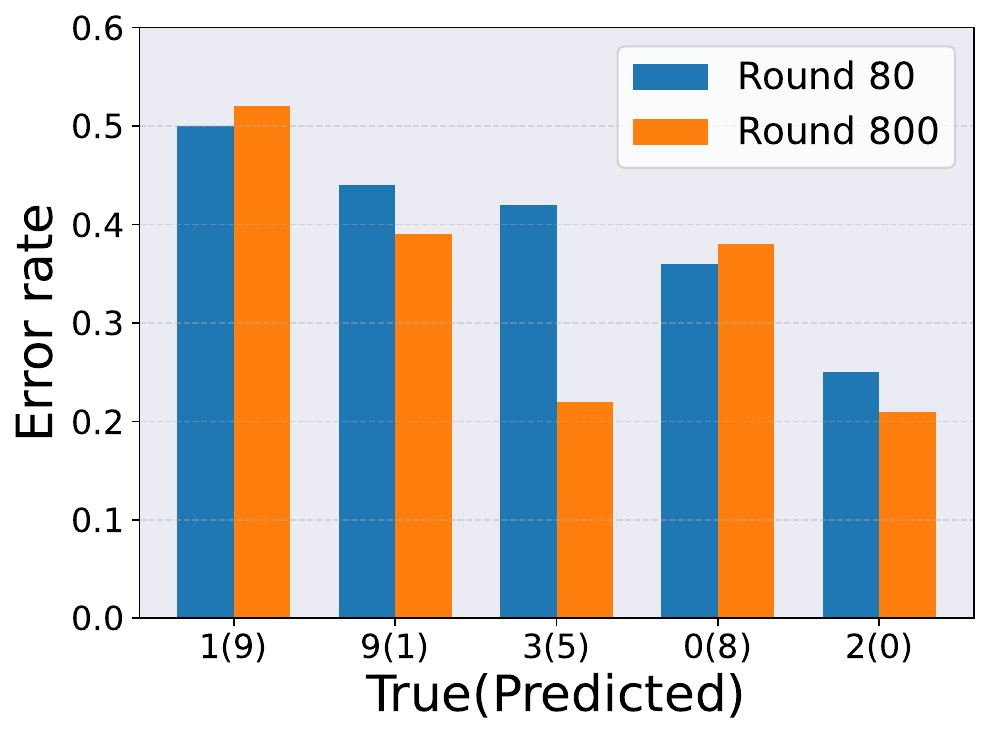}
    \caption{SemiFL}
    \label{fig:semifl_top5}
  \end{subfigure}
  \hfill
  \begin{subfigure}[b]{0.32\linewidth}
    \centering
    \includegraphics[width=\linewidth]{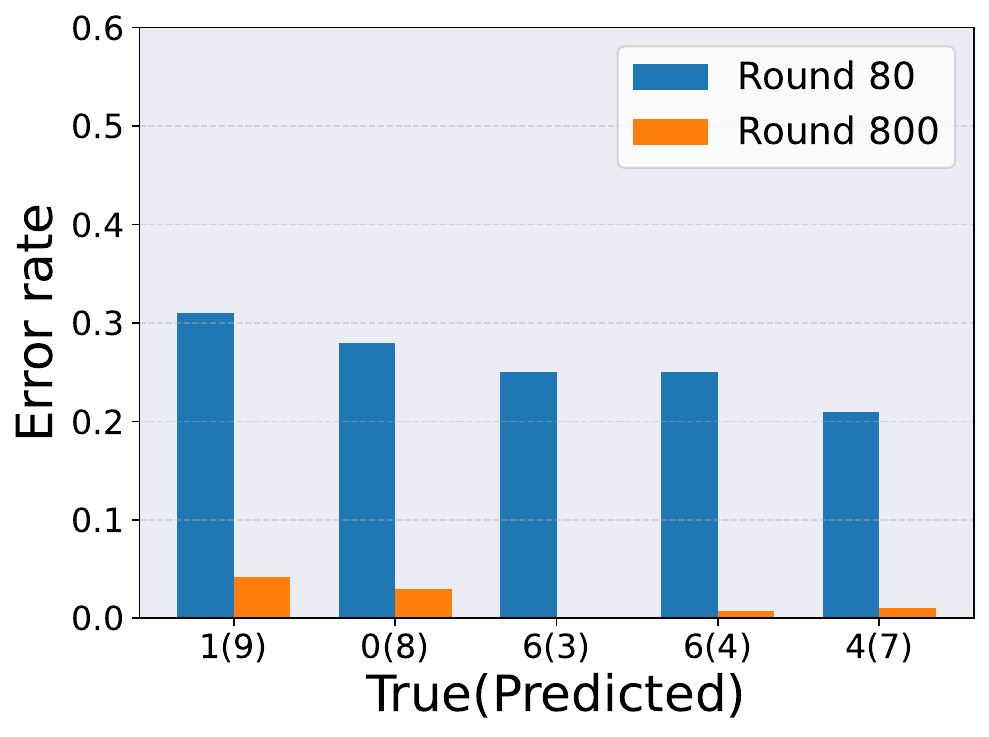}
    \caption{CATCHFed}
    \label{fig:catchfed_top5}
  \end{subfigure}
  \caption{(a) Pseudo-label confusion matrices at rounds 80 and 800, (b) SemiFL, (c) CATCHFed comparison of the five most frequent mis-classified class pairs (CIFAR-10, IID, 40 labels).}
  \label{fig:conf_mat_comparison}
  \vspace{-5pt}
\end{figure*}

\begin{table*}[!b]
\vspace{-5pt}
  \centering
  \begin{minipage}[t]{0.49\linewidth}
    \centering
    \caption{Ablation study of CATCHFed on CIFAR-10, IID 40 labels. Unpseudo: consistency regularization with unpseudo-labeled data.}
    \resizebox{\linewidth}{!}{
    \begin{tabular}{c c c c c c}
      \toprule
      \textbf{Exp} & \textbf{CAWT} & \textbf{Unpseudo} & \textbf{Hybrid} & \textbf{Acc} & \textbf{PL Acc} \\
      \midrule
      1 & \multicolumn{3}{c}{Baseline(SemiFL)} & 65.19 & 68.22 \\
      \midrule
      2 & \ding{51} & & & 73.09 & 73.63 \\
      3 & & \ding{51} & & 71.3 & 76.11 \\
      4 & \ding{51} & \ding{51} & & 73.04 & 73.88 \\
      \rowcolor{yellow!50}5 & \ding{51} & \ding{51} & \ding{51} & \textbf{77.45} & \textbf{81.86} \\
      \bottomrule
    \end{tabular}}
    \label{tab:iid40}
  \end{minipage}
  \hfill
  \begin{minipage}[t]{0.49\linewidth}
    \centering
    \caption{Ablation study of CATCHFed on CIFAR-10, IID 20 labels. Unpseudo: consistency regularization with unpseudo-labeled data.}
    \resizebox{\linewidth}{!}{%
    \begin{tabular}{c c c c c c}
      \toprule
      \textbf{Exp} & \textbf{CAWT} & \textbf{Unpseudo} & \textbf{Hybrid} & \textbf{Acc} & \textbf{PL Acc} \\
      \midrule
      1 & \multicolumn{3}{c}{Baseline(SemiFL)} & 35.11 & 36.07 \\
      \midrule
      2 & \ding{51} & & & 47.34 & 47.65 \\
      3 & & \ding{51} & & 42.94 & 45.25 \\
      \rowcolor{yellow!50}4 & \ding{51} & \ding{51} & & \textbf{49.02} & \textbf{49.33} \\
      5 & \ding{51} & \ding{51} & \ding{51} & 46.33 & 48.81 \\
      \bottomrule
    \end{tabular}}
    \label{tab:iid20}
  \end{minipage}
  \vspace{-5pt}
\end{table*}

\subsection{Effectiveness of CAWT} \label{sec:cawt_effect}

\subsubsection{Classwise performance}\quad The effectiveness of CAWT’s adaptive threshold in reflecting class-aware learning difficulties can be confirmed by examining classwise test accuracy. As shown in Fig. \ref{fig:classwise_acc_compare}, the existing methods exhibited relatively poor performance on certain classes (e.g., Classes 2, 3, and 6), whereas CATCHFed consistently achieved higher accuracy across all classes. These results suggest that CAWT, which considers class-wise learning difficulty for each client, can provide proper thresholds for classes with different confidences.

\begin{figure*}[t]
\vspace{-5pt}
  \centering
  \begin{subfigure}[b]{0.32\linewidth}
    \centering
    \includegraphics[width=\linewidth]{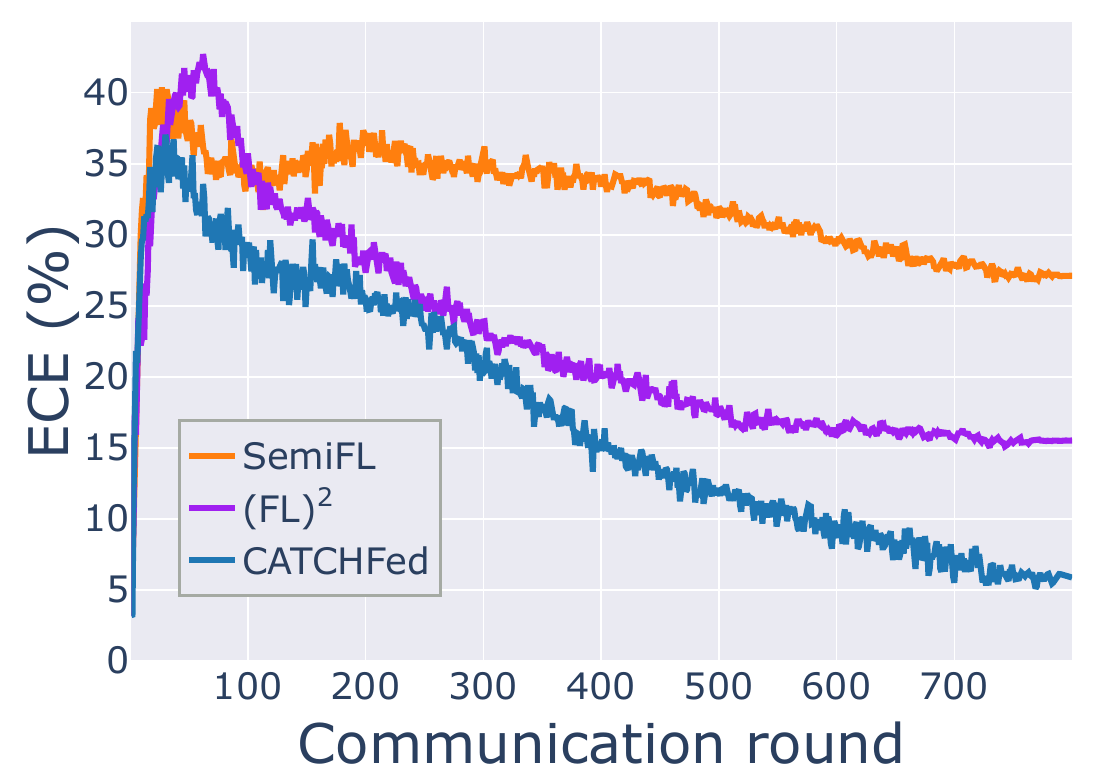}
    \caption{ECE graph}
    \label{fig:ece_graph}
  \end{subfigure}
  \hfill
  \begin{subfigure}[b]{0.22\linewidth}
    \centering
    \includegraphics[width=\linewidth]{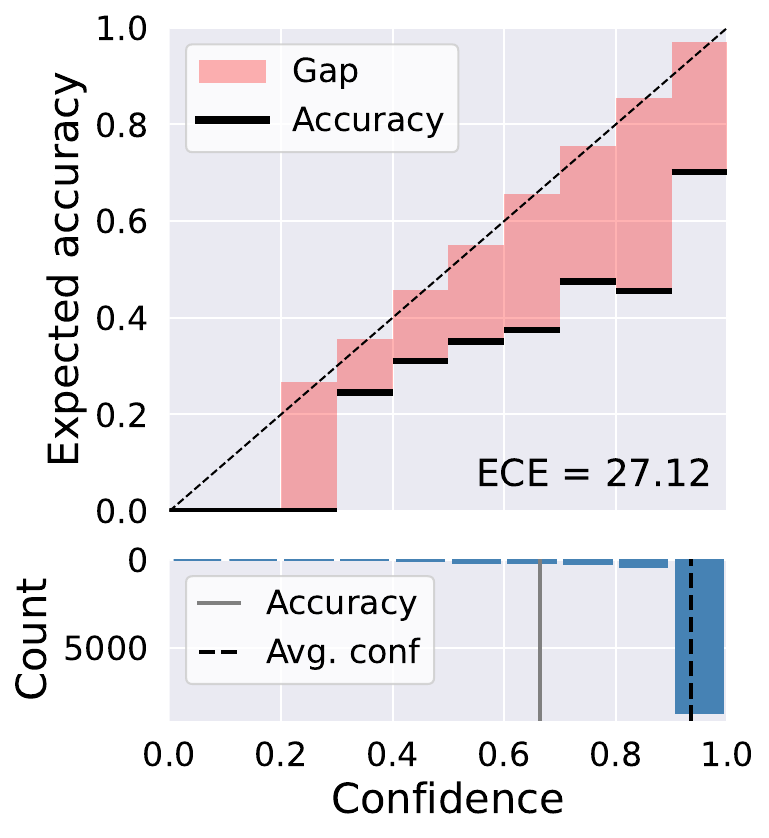}
    \caption{SemiFL}
    \label{fig:semifl_rd}
  \end{subfigure}
  \hfill
  \begin{subfigure}[b]{0.22\linewidth}
    \centering
    \includegraphics[width=\linewidth]{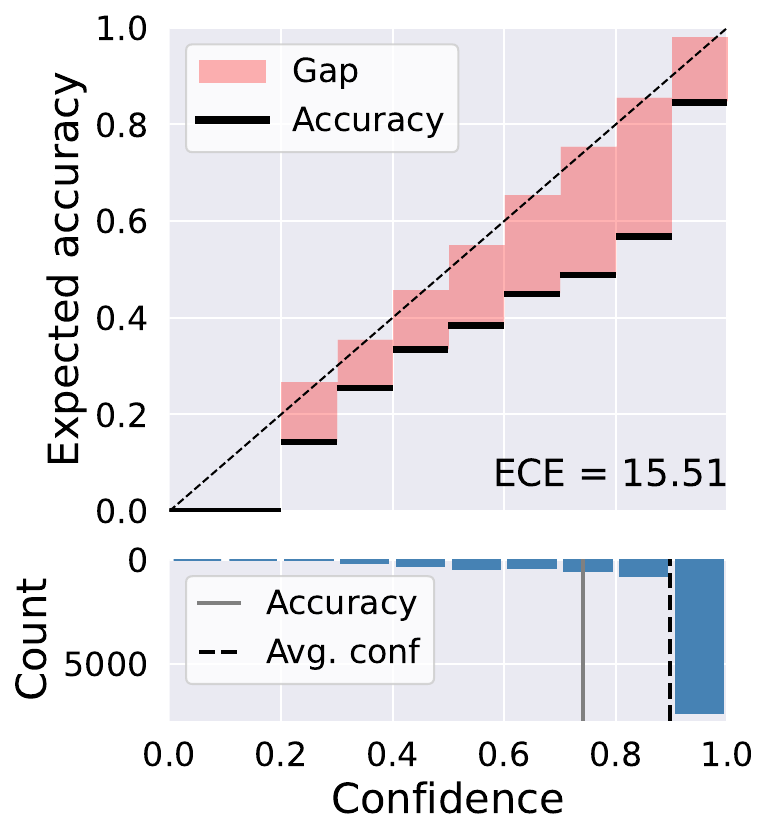}
    \caption{$(FL)^2$}
    \label{fig:flfl_rd}
  \end{subfigure}
  \hfill
  \begin{subfigure}[b]{0.22\linewidth}
    \centering
    \includegraphics[width=\linewidth]{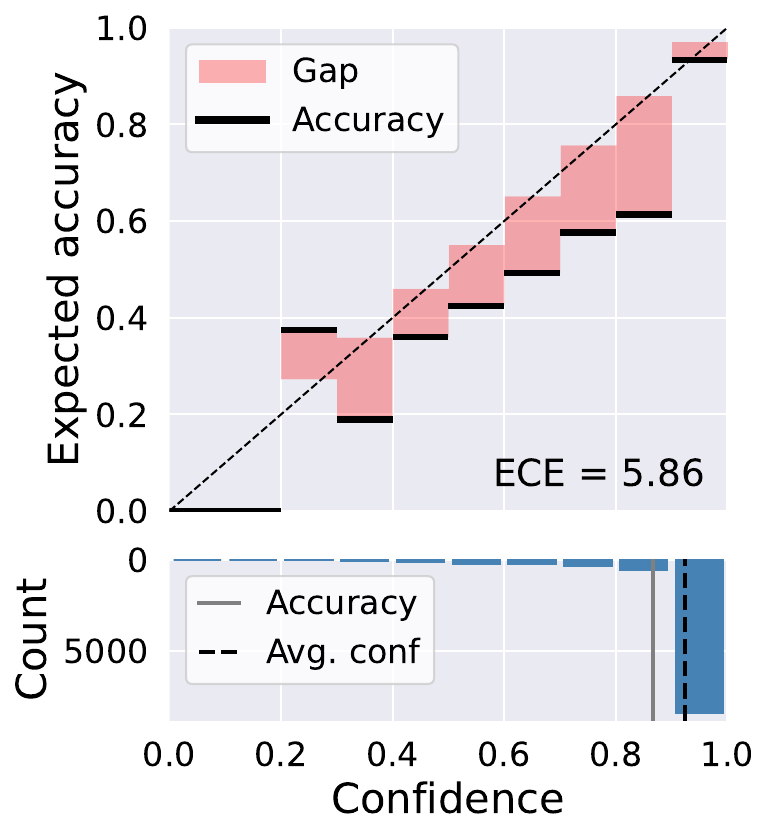}
    \caption{CATCHFed}
    \label{fig:catchfed_rd}
  \end{subfigure}
  \caption{(a) ECE over communication rounds. (b)--(d) Reliability diagrams (top) and confidence histograms (bottom) for SemiFL, $(FL)^2$, and CATCHFed at round 800. (CIFAR-10, IID 40 labels).}
  \label{fig:ece_compare}
  \vspace{-5pt}
\end{figure*}

\subsubsection{Effectiveness of warm-up phase}\quad We analyzed pseudo-label confusion matrices to determine whether CAWT's warm-up phase effectively mitigates early confirmation bias. As shown in Fig. \ref{fig:pl_conf_mat}, both SemiFL and CATCHFed initially assigned incorrect pseudo-labels concentrated on certain classes. However, while SemiFL \cite{diao2022semifl} reinforced these incorrect predictions over training, CATCHFed gradually reduced them. This difference is quantitatively illustrated in Figs. \ref{fig:semifl_top5} and \ref{fig:catchfed_top5}, where CATCHFed consistently maintained lower pseudo-label error rates than SemiFL from early training stages. These results indicate that the warm-up phase enabled more unlabeled data to participate effectively, reducing early confirmation bias toward particular classes. This observation aligns closely with our expectations regarding the effectiveness of CAWT.

\begin{figure}[t]
    \centering
    \begin{subfigure}[t]{0.48\columnwidth}
        \centering
        \includegraphics[width=\linewidth]{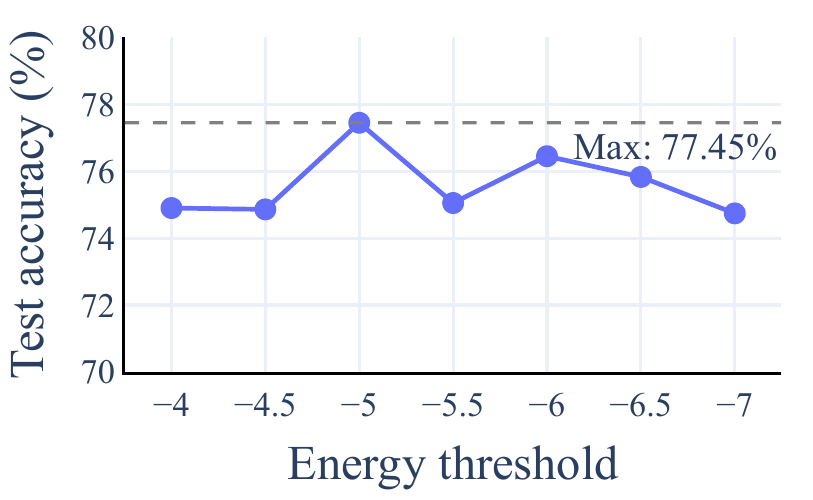}
        \caption{}
        \label{fig:energy_score_compare}
    \end{subfigure}
    \hfill
    \begin{subfigure}[t]{0.48\columnwidth}
        \centering
        \includegraphics[width=\linewidth]{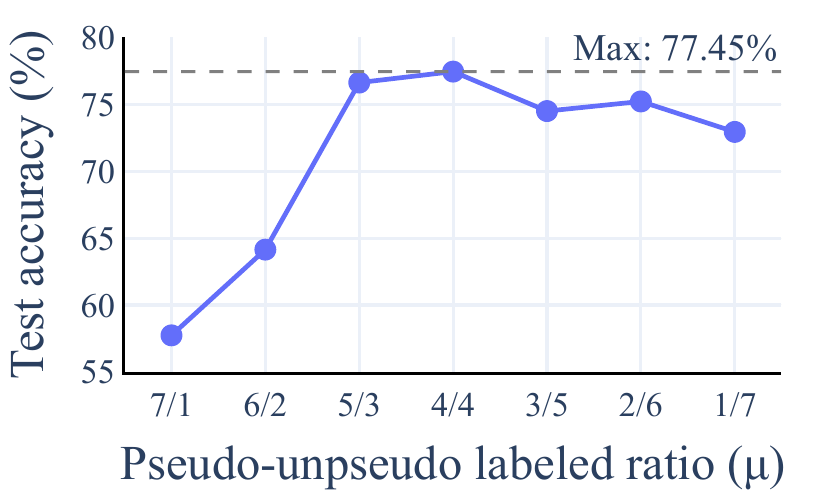}
        \caption{}
        \label{fig:ratio_compare}
    \end{subfigure}
    \vspace{-5pt}
    \caption{
        (a) Effect of energy threshold on test accuracy. 
        (b) Test accuracy across different pseudo and unpseudo-labeled ratios. 
        (CIFAR-10, IID, 40 labels)
    }
    \label{fig:energy_and_ratio}
\end{figure}

\subsection{Ablation study} \label{sec:ablation_study}

We conducted an ablation study to analyze the contribution of each component of CATCHFed to overall performance.
Experiments were performed across environments with varying numbers of labeled data, with results presented in Tables \ref{tab:iid40} and \ref{tab:iid20}.
We adopt SemiFL as the baseline method. Experimental results consistently show that applying each individual component separately
leads to superior performance compared to the baseline across all tested environments.
In some cases (Exp5: CATCHFed), full CATCHFed results in lower performance compared to scenarios with fewer combined components as shown in Table \ref{tab:iid20}. We attribute this degradation to hyperparameter tuning issues. Specifically, as depicted in Fig. \ref{fig:energy_score_compare}, the Energy threshold tuning was optimized under the CIFAR-10 IID 40 setting, whereas the Energy score is known to be sensitive to model architectures and dataset characteristics \cite{yu2023inpl}. It implies that the sub-optimal threshold values may degrade the overall performance in different settings. Further discussion and improvements regarding the energy threshold tuning are provided in Section \ref{sec:energy_threshold_setting}.
Similarly, the reason Exp4 shows no additional improvement compared to Exp2 in Table \ref{tab:iid40} can also be attributed to the setting of the ratio $\mu$. As shown in Fig. \ref{fig:ratio_compare}, the test accuracy varies significantly depending on the pseudo-to-unpseudo-labeled ratio, indicating that the choice of $\mu$ directly affects the effectiveness of each component.

\subsection{Energy threshold setting} \label{sec:energy_threshold_setting}
Since the choice of energy threshold influences model performance across different environments, we conducted experiments to examine how the optimal value varies across datasets. As shown in Table \ref{tab:energy_threshold_setting}, performance on CIFAR-10, CIFAR-100, and SVHN changes noticeably with different thresholds, indicating that the model is sensitive to this parameter. This variation appears to be related to two dataset-specific characteristics: classification difficulty and the number of classes. In relatively simple tasks, the model captures the underlying data distribution more effectively, resulting in lower energy scores for in-distribution samples, where a stricter (i.e., lower) threshold tends to improve performance. Moreover, as indicated in Eq. \ref{eq:energy_score}, an increase in the number of classes increases the number of summed logit terms inside the logarithm, thereby reducing the average energy score. Consequently, datasets with more classes generally require a lower optimal threshold. These results suggest that, although threshold selection affects performance, an appropriate value can be estimated by considering dataset characteristics such as classification difficulty and class count.

\begin{table}[t]
\centering
\caption{Test accuracy (\%) across different energy thresholds for different datasets.}
\label{tab:energy_threshold_setting}
\resizebox{\linewidth}{!}{
\begin{tabular}{c c c c c c c c}
\toprule
\textbf{Threshold} & \textbf{-4.0} & \textbf{-4.5} & \textbf{-5.0} & \textbf{-5.5} & \textbf{-6.0} & \textbf{-6.5} & \textbf{-7.0} \\
\midrule
\textbf{CIFAR-10}   & 74.90 & 74.86 & \textbf{77.45} & 75.05 & 76.45 & 75.83 & 74.74 \\
\textbf{CIFAR-100}  & 44.06 & 44.06 & 44.59 & 45.03 & 44.53 & \textbf{45.13} & 43.64 \\
\textbf{SVHN}       & 82.95 & 84.07 & 82.38 & 85.89 & 85.35 & 86.65 & \textbf{87.30} \\
\bottomrule
\end{tabular}
}
\end{table}

\subsection{Calibration of CATCHFed}

The confirmation bias can be quantified with calibration metrics \cite{chen2022semi}. 
We evaluate the confirmation bias of CATCHFed using standard calibration indicators such as Expected Calibration Error (ECE), confidence histograms, and reliability diagrams, as illustrated in Fig. \ref{fig:ece_compare}. CATCHFed consistently reduces ECE over training rounds, improving model calibration. Specifically, Fig.~\ref{fig:semifl_rd} shows that SemiFL yields a high ECE of $27.12$, indicating systematic over-confidence in its probability predictions. Fig. \ref{fig:flfl_rd} shows that $(FL)^2$ achieves a lower ECE of $15.5$1 compared to SemiFL, suggesting partial mitigation of confirmation bias, though some degree of over-confidence remains. Conversely, Fig. \ref{fig:catchfed_rd} demonstrates that CATCHFed achieves a low ECE of $5.86$, indicating a close match between predicted probabilities and actual accuracies, i.e. less confirmation bias. Particularly at high confidence level, CATCHFed effectively mitigates confirmation bias and demonstrates strong calibration, implying that pseudo-labels generated with high prediction confidence are also highly reliable.

\section{Conclusion}

In this paper, we propose three methods to efficiently utilize unlabeled client data in SSFL environments with extremely limited labeled data. First, to enhance the accuracy and usage rate of pseudo-labels, we introduce CAWT, which adaptively adjusts thresholds based on client-specific learning difficulty, and Hybrid Thresholding, which jointly considers model confidence and whether data is in-distribution. Additionally, we leverage discarded unpseudo-labeled samples from threshold-based methods for consistency regularization, further enhancing the utilization of unlabeled data. Extensive experiments confirm that our methods significantly outperform existing techniques, particularly in label-scarce settings.
Moreover, we empirically verified that it is possible to infer an approximate optimal range for the energy threshold, which plays a critical role in pseudo-label quality and overall performance.
Future work will focus on analyzing the theoretical convergence properties of the proposed methods and extending their applicability to a wider range of learning scenarios.

\bibliographystyle{IEEEtran}
\bibliography{ref}

\end{document}